\documentclass[11pt, a4paper, logo, copyright, nonumbering]{map}

\usepackage[justification=justified]{caption} 

\definecolor{hidden-draw}{RGB}{20,68,106}
\definecolor{hidden-pink}{RGB}{255,245,247}
\usepackage{natbib}

\usepackage{CJKutf8}

\usepackage{xargs}  

\usepackage{todonotes}  

\usepackage{multirow}

\usepackage{cleveref}

\usepackage{amsmath}
\usepackage{dsfont}

\usepackage{subcaption}

\usepackage{svg}
 \usepackage{ragged2e} 
\usepackage{tikz}
\usepackage{threeparttable}   
\usepackage{multicol}

\usetikzlibrary{arrows.meta, positioning, calc}

\usepackage{graphicx}

\usepackage[edges]{forest}

\definecolor{bred}{RGB}{250, 82, 82}
\definecolor{borange}{RGB}{253, 126, 20}
\definecolor{byellow}{RGB}{250, 176, 5}
\definecolor{bgreen}{RGB}{116, 184, 22}
\definecolor{bblue}{RGB}{250, 176, 5}
\definecolor{bindigo}{RGB}{76, 110, 245}
\definecolor{bcyan}{RGB}{59, 201, 219}
\definecolor{bteal}{RGB}{99, 230, 190}

\usepackage{amsmath,amsfonts,bm}

\def\eqref#1{equation~\ref{#1}}

\def\1{\bm{1}}

\DeclareMathAlphabet{\mathsfit}{\encodingdefault}{\sfdefault}{m}{sl}
\SetMathAlphabet{\mathsfit}{bold}{\encodingdefault}{\sfdefault}{bx}{n}

\usepackage{float}   

\hypersetup{
    colorlinks=true,            
    linkcolor=blue,             
    filecolor=magenta,          
    urlcolor=cyan,              
    citecolor=purple,             
    pdftitle={Latent Survey},
    pdfpagemode=FullScreen,
    }

\renewcommand{\today}{}
\newcommandx{\info}[2][1=]{\todo[linecolor=red,backgroundcolor=red!25,bordercolor=red,#1]{#2}}

\title{\centering A Survey on Latent Reasoning}

\author{
\small
\textbf{Rui-Jie Zhu$^{\star,\dagger}$, 
Tianhao Peng$^{\star}$, 
Tianhao Cheng$^{\star}$,
Xingwei Qu$^{\star}$,} \\
\small
Jinfa Huang,
Dawei Zhu,
Hao Wang,
Kaiwen Xue,
Xuanliang Zhang,
Yong Shan,
Tianle Cai,
Taylor Kergan,
Assel Kembay,
Andrew Smith,
Chenghua Lin,
Binh Nguyen,
Yuqi Pan,
Yuhong Chou,\\
\small
Zefan Cai,
Zhenhe Wu,
Yongchi Zhao,
Tianyu Liu,
Jian Yang,
Wangchunshu Zhou,\\
\small
Chujie Zheng,
Chongxuan Li,
Yuyin Zhou,
Zhoujun Li,
Zhaoxiang Zhang, \\
\small
Jiaheng Liu$^{\dagger}$,
Ge Zhang$^{\dagger}$,
Wenhao Huang,
Jason Eshraghian$^{\dagger}$
\\
        \vspace{0.1in}
UCSC, FDU, NJU, PKU, RUC, UoM, UW-Madison, PolyU, M-A-P
    \vspace{-0.35in}

}

\vspace{-0.35in}

\begin{abstract}
    \vspace{-0.15in}
Large Language Models (LLMs) have demonstrated impressive reasoning capabilities, especially when guided by explicit chain-of-thought (CoT) reasoning that verbalizes intermediate steps. While CoT improves both interpretability and accuracy, its dependence on natural language reasoning limits the model’s expressive bandwidth. Latent reasoning tackles this bottleneck by performing multi-step inference entirely in the model’s continuous hidden state, eliminating token-level supervision. To advance latent reasoning research, this survey provides a comprehensive overview of the emerging field of latent reasoning. We begin by examining the foundational role of neural network layers as the computational substrate for reasoning, highlighting how hierarchical representations support complex transformations. Next, we explore diverse latent reasoning methodologies, including activation-based recurrence, hidden state propagation, and fine-tuning strategies that compress or internalize explicit reasoning traces. Finally, we discuss advanced paradigms such as infinite-depth latent reasoning via masked diffusion models, which enable globally consistent and reversible reasoning processes. By unifying these perspectives, we aim to clarify the conceptual landscape of latent reasoning and chart future directions for research at the frontier of LLM cognition. An associated GitHub repository collecting the latest papers and repos is available at: \href{https://github.com/multimodal-art-projection/LatentCoT-Horizon/}{\color[RGB]{175,36,67}{LatentCoT-Horizon}}.
\end{abstract}

\begin{document}
\maketitle

\let\oldthefootnote\thefootnote

\let\thefootnote\relax\footnotetext{$^\star$~Equal Contribution. ~~$^\dagger$~Corresponding Authors. }
\let\thefootnote\oldthefootnote

\begin{figure}[H] 
    \vspace{-2mm}

    \centering    \includegraphics[width=0.8\textwidth]{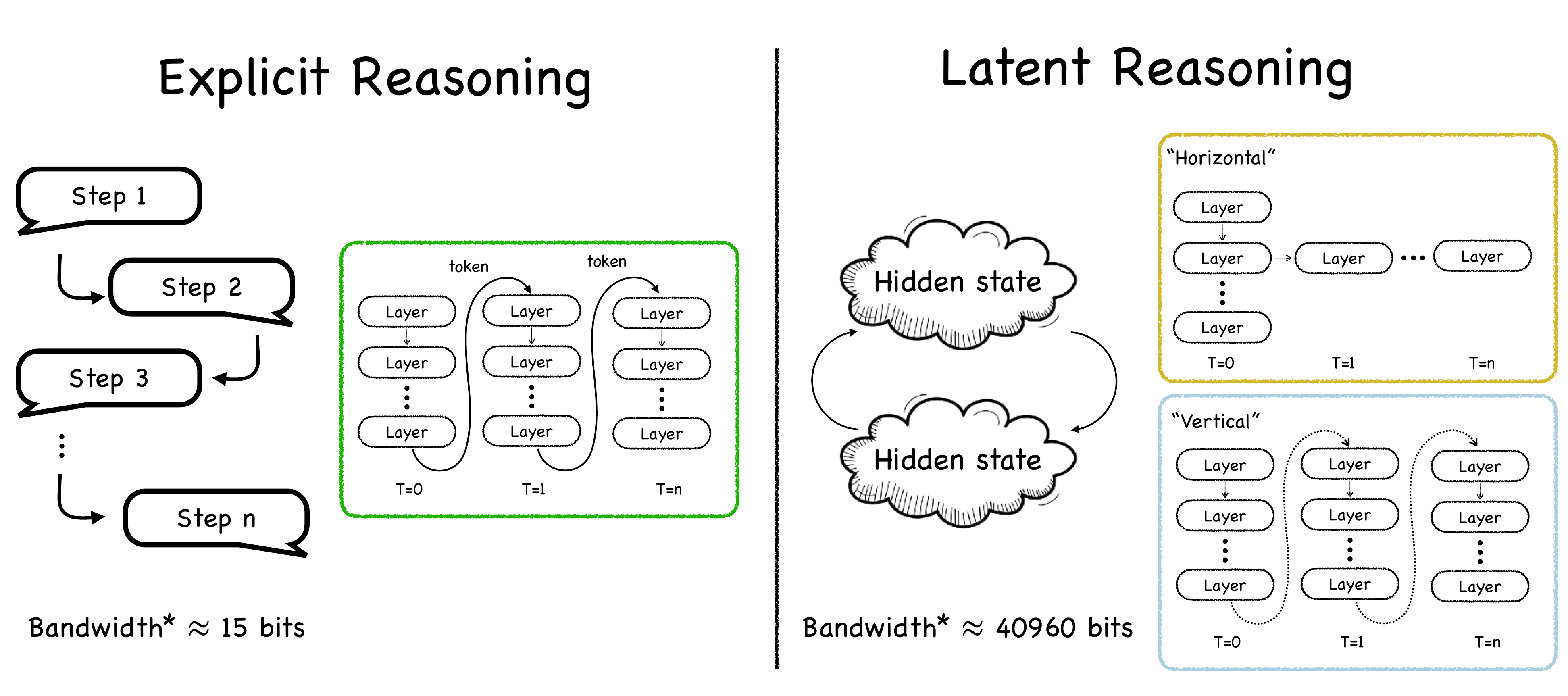}
        \vspace{-2mm}
    \caption{Explicit reasoning transmits discrete tokens ($\approx 15\,\text{bits}$ each), whereas latent reasoning exchanges full 2560-dimensional FP16 hidden states ($\approx 40,960\,\text{bits}$ each), revealing a $\sim\!2.7\times10^{3}$-fold bandwidth gap between the two approaches.}
    \label{fig:concept}
\end{figure}

\newpage

\tableofcontents

\newpage

\section{Introduction}
Large Language Models (LLMs) have demonstrated remarkable capabilities in performing reasoning tasks, in some cases even exceeding human-level performance~\citep{liu2025comprehensive,zhang2024mapneo,Huang2024OpenCoderTO,que2024dcpt}. LLMs often reason more effectively when they produce a Chain-of-Thought (CoT)~\citep{wei2022chain}, spelling out each intermediate step in natural language before arriving at a final answer. 

Initially viewed as a logical extension to prompt engineering, CoT gained traction once supervised instruction tuning exposed models to many annotated reasoning traces. It then became the norm when RL rewarded answer correctness~\citep{jaech2024o1}, which encouraged models to generate their own effective chains of thought. As a result, LLMs that ``think in language before answering” have attained remarkable performance improvements. This principle now anchors leading reasoning models, including the Qwen3 series~\citep{yang2025qwen3}, DeepSeek-R1~\citep{guo2025deepseek}, and Gemini 2.5 series~\citep{google2025gemini2.5}.

However, just as humans do not always rely on language for their cognitive processes,
LLMs spend most of their processing budget in the latent space. Enforcing a CoT to
operate with natural language can constrain a model’s expressive range and can also
impose redundant computation. Latent Chain-of-Thought (Latent CoT) has the potential
to overcome these limits~\citep{deng2024explicit, hao2024training}. Unlike its explicit
counterpart that relies on discrete tokens, latent CoT carries reasoning in continuous
internal representations, often via recurrent mechanisms within the model. This offers
richer expressivity and access to non-linguistic reasoning paths, potentially unlocking
new frontiers in model reasoning.

This survey examines the emerging landscape of Latent CoT and its potential to
surpass language-based reasoning constraints. While explicit CoT forces thoughts into
a string of tokens, Latent CoT shifts the entire reasoning process into the model’s
continuous representational space. The aim is to expand expressiveness and raise the
performance ceiling: freed from a finite vocabulary, a model can explore reasoning
trajectories with no direct linguistic equivalent. We categorize and analyze the technical
approaches that leverage these continuous representations to achieve more advanced
reasoning.

The structure of this survey is designed to provide a comprehensive understanding of
Latent CoT and its various implementations. Our taxonomy breaks this down in
Figure~\ref{fig:taxo_of_lr}. We begin by establishing a general formulation that captures
most Latent CoT implementations, before classing techniques into more specific
categories. These categories can be broadly divided into two types: 1) \textbf{vertical
recurrence} for expanding computational depth, and 2) \textbf{horizontal recurrence} for
increasing sequential capacity. Vertical recurrence applies feedback loops to activation
values, and can be thought of `activation-based’
reasoning~\cite{dehghani2018universal, mohtashami2023cotformer}. Alternatively,
horizontal recurrence uses hidden states to propagate context across long sequences
of reasoning trajectories~\citep{sun2024ttt, schone2025implicit}. We then explore fine-tuning strategies designed to compress or internalize explicit reasoning traces, which
concludes the review of Latent CoT implementations.

This sets the stage for understanding the mechanistic interpretability of latent reasoning
to understand how these processes are realized within neural networks. This section
examines the foundational role of network layers as the primary computational substrate
for reasoning~\citep{zhang2024investigating, shi2024understanding}. We explore the
theory of Layer Specialization, which posits that different layers develop distinct,
hierarchical functions—from feature extraction in shallow layers to complex logical
operations in intermediate layers and final integration in deep layers—collectively
forming an implicit computational pipeline analogous to an explicit CoT. Explicit CoT
comes with the benefit of intermediate tokens which offers a degree of post-hoc interpretability, and we similarly aim to uncover the mechanisms that enable latent
reasoning.

Finally, we explore advanced paradigms at the frontier of LLM cognition, focusing on the
pursuit of infinite-depth reasoning. This concept refers to a model's ability to devote
unbounded computational steps to refine a solution, moving beyond fixed-depth
architectures. Our discussion centers on spatial infinite reasoning as realized by text
diffusion models~\citep{ye2024diffusionthoughtschainofthoughtreasoning,
nie2024scaling}. Unlike traditional autoregressive generation, these models operate on
the entire output sequence in parallel, enabling global planning and iterative self-
correction through bidirectional context. This approach facilitates globally consistent and
reversible reasoning processes, offering a promising path toward more powerful and
flexible AI systems.
\tikzstyle{my-box}=[
    rectangle,
    draw=hidden-draw,
    rounded corners,
    text opacity=1,
    minimum height=1.5em,
    minimum width=5em,
    inner sep=2pt,
    align=center,
    fill opacity=.5,
    line width=0.8pt,
]
\tikzstyle{leaf}=[my-box, minimum height=1.5em,
    fill=white, text=black, align=left,font=\normalsize,
    inner xsep=2pt,
    inner ysep=4pt,
    line width=0.8pt,
]

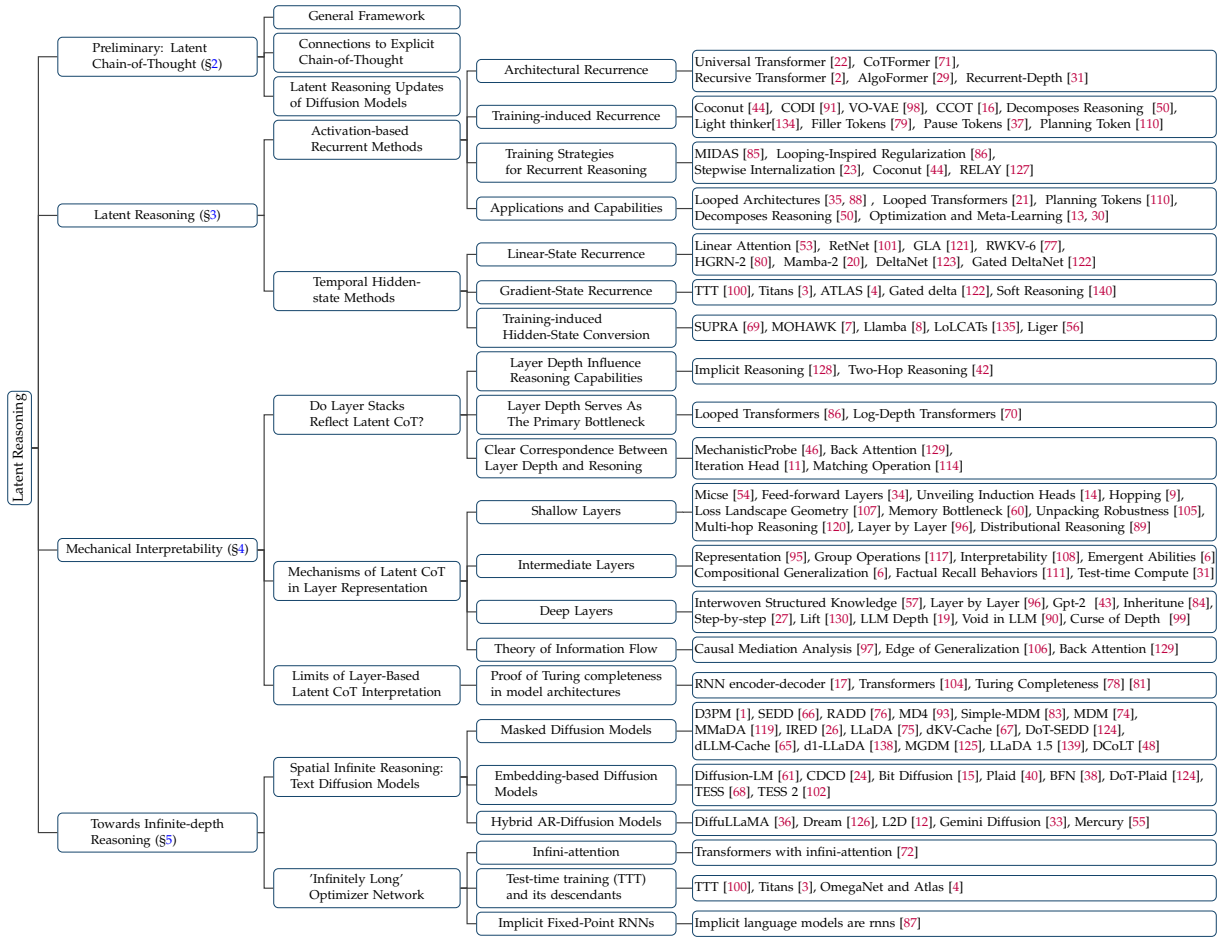
\begin{figure}[t!]
    \centering
    \resizebox{\textwidth}{!}{
        \begin{forest}
        forked edges,
            for tree={
                grow=east,
                reversed=true,
                anchor=base west,
                parent anchor=east,
                child anchor=west,
                base=center,
                font=\large,
                rectangle,
                draw=hidden-draw,
                rounded corners,
                align=left,
                text centered,
                minimum width=4em,
                edge+={darkgray, line width=1pt},
                s sep=3pt,
                inner xsep=2pt,
                inner ysep=3pt,
                line width=0.8pt,
                ver/.style={rotate=90, child anchor=north, parent anchor=south, anchor=center}
            },
            where level=1{text width=15em,font=\normalsize}{},
            where level=2{text width=14em,font=\normalsize}{},
            where level=3{text width=15em,font=\normalsize}{},
            where level=4{text width=15em,font=\normalsize}{},
            where level=5{text width=15em,font=\normalsize}{},
            [
                Latent Reasoning, ver
                [
                    Preliminary: Latent \\Chain-of-Thought (\S\ref{sec:preliminary})
                    [
                        General Framework
                    ]
                    [
                        Connections to Explicit \\Chain-of-Thought
                    ]
                    [
                        Latent Reasoning Updates \\of Diffusion Models
                    ]
                ]
                [
                    Latent Reasoning (\S\ref{sec:latent cot reasoning})
                    [
                        Activation-based \\Recurrent Methods
                        [
                            Architectural Recurrence
                            [
                                Universal Transformer~\citep{dehghani2018universal}{, }
                                CoTFormer~\citep{mohtashami2023cotformer}{, }\\
                                Recursive Transformer~\citep{bae2024relaxed}{, }
                                AlgoFormer~\citep{gao2024algoformer}{, }
                                Recurrent-Depth~\citep{geiping2025scaling}
                                , leaf,text width=40em
                            ]
                        ]
                        [
                            Training-induced Recurrence
                            [
                                Coconut~\citep{hao2024training}{, }
                                CODI~\citep{shen2025codi}{, }VO-VAE~\citep{su2025token}{, }
                                CCOT~\citep{cheng2024compressed}{, }Decomposes Reasoning ~\citep{jin2024disentangling}{, }
                                \\Light thinker\citep{zhang2025lightthinker}{, }
                                Filler Tokens~\citep{pfau2024lets}{, }
                                Pause Tokens~\citep{goyal2024think}{, }
                                Planning Token~\citep{wang2024guiding}
                                , leaf,text width=40em
                            ]    
                        ]
                        [
                            Training Strategies \\for Recurrent Reasoning
                            [
                                MIDAS~\citep{saunshi2024inductive}{, }
                                Looping-Inspired Regularization~\citep{saunshi2025reasoning}{, }\\
                                Stepwise Internalization~\citep{deng2024explicit}{, }
                                Coconut~\citep{hao2024training}{, }
                                RELAY~\citep{yu2025enhancing}
                                , leaf,text width=40em
                            ]
                        ]
                        [
                            Applications and Capabilities
                            [
                                Looped Architectures~\citep{schwarzschild2021can,giannou2023looped} {, }
                                Looped Transformers~\citep{de2024simulation}{, }
                                Planning Tokens~\citep{wang2024guiding}{, }\\
                                Decomposes Reasoning~\citep{jin2024disentangling}{, }
                                Optimization and Meta-Learning~\citep{gatmiry2024can, chen2024bypassing}
                                , leaf,text width=40em
                            ]
                        ]
                    ]
                    [
                        Temporal Hidden-\\state Methods
                        [
                            Linear-State Recurrence
                            [
                                Linear Attention~\citep{katharopoulos2020transformers}{, }
                                RetNet~\citep{sun2023retentive}{, }
                                GLA~\citep{yang2023gated}{, }
                                RWKV-6~\citep{peng2024eagle}{, }\\
                                HGRN-2~\citep{qin2024hgrn2}{, }
                                Mamba-2~\citep{dao2024transformers}{, }
                                DeltaNet~\citep{yang2024deltanet}{, }
                                Gated DeltaNet~\citep{yang2024gateddeltanet}
                                , leaf,text width=40em
                            ]
                        ]
                        [
                            Gradient-State Recurrence
                            [
                                TTT~\citep{sun2024ttt}{, }Titans~\citep{behrouz2024titans}{, }ATLAS~\citep{behrouz2025atlas}{, }Gated delta~\citep{yang2024gateddeltanet}{, }Soft Reasoning~\cite{zhu2025soft}
                                , leaf,text width=40em
                            ]
                        ]
                        [
                            Training-induced \\Hidden-State Conversion
                            [
                                SUPRA~\citep{mercat2024supra}{, }MOHAWK~\citep{bick2024mohawk}{, }Llamba~\citep{bick2025llamba}{, }LoLCATs~\citep{zhang2024lolcats}{, }Liger~\citep{lan2025liger}
                                , leaf,text width=40em
                            ]
                        ]
                    ]
                ]
                [
                    Mechanical Interpretability (\S\ref{sec:interpretability})
                    [
                        Do Layer Stacks \\Reflect Latent CoT?
                        [
                            Layer Depth Influence \\Reasoning Capabilities
                            [
                            Implicit Reasoning~\citep{yu2024llms}{, } 
                            Two-Hop Reasoning~\citep{guo2025llms}
                            , leaf,text width=40em
                            ]
                        ]
                        [
                            Layer Depth Serves As \\The Primary Bottleneck
                            [
                            Looped Transformers~\citep{saunshi2025reasoning}{,} Log-Depth Transformers~\citep{merrill2025little}
                            , leaf,text width=40em
                            ]
                        ]
                        [
                            Clear Correspondence Between \\Layer Depth and Resoning
                            [
                             MechanisticProbe~\citep{hou2023towards}{, }Back Attention~\citep{yu2025back}{,} \\Iteration Head~\citep{cabannes2024iteration}{,} 
                            Matching Operation~\citep{wang2024towards}
                            , leaf,text width=40em
                            ]
                        ]
                    ]
                    [
                        Mechanisms of Latent CoT \\in Layer Representation
                        [
                            Shallow Layers
                            [
                            Micse~\citep{klein2022micse}{,} Feed-forward Layers~\citep{geva2020transformer}{,} Unveiling Induction Heads~\citep{chen2024unveiling}{,}  Hopping~\citep{biran2024hopping}{,} \\
                            Loss Landscape Geometry~\citep{wang2024loss}{,} Memory Bottleneck~\citep{li2019enhancing}{,} Unpacking Robustness~\citep{walkowiak2025unpacking}{,} \\
                            Multi-hop Reasoning~\citep{yang2024large}{,} Layer by Layer~\citep{skean2025layer}{,} 
                            Distributional Reasoning~\citep{shalev2024distributional}
                            , leaf,text width=40em
                            ]
                        ]
                        [
                            Intermediate Layers
                            [
                            Representation~\citep{skean2024does}{,} Group Operations~\citep{wu2024unifying}{,} Interpretability~\citep{wang2022interpretability}{,} Emergent Abilities~\citep{berti2025emergent}\\Compositional Generalization~\citep{berti2025emergent}{,} 
                            Factual Recall Behaviors~\citep{wang2024unveiling}{,} Test-time Compute~\citep{geiping2025scaling}
                            , leaf,text width=40em
                            ]
                        ]
                        [
                            Deep Layers
                            [
                            Interwoven Structured Knowledge~\citep{lei2025representation}{,} Layer by Layer~\citep{skean2025layer}{,} Gpt-2 ~\citep{hanna2023does}{,} Inheritune~\citep{sanyal2024inheritune}{,} \\Step-by-step~\citep{dutta2024think}{,} Lift~\citep{yuan2024lift}{,} LLM Depth~\citep{csordas2025language}{,} Void in LLM~\citep{shemiranifar2025void}{,} Curse of Depth ~\citep{sun2025curse}
                            , leaf,text width=40em
                            ]
                        ]
                        [
                            Theory of Information Flow
                            [
                            Causal Mediation Analysis~\citep{stolfo2023mechanistic}{,} Edge of Generalization~\citep{wang2024grokked}{,} Back Attention~\citep{yu2025back}
                            , leaf,text width=40em
                            ]
                        ]
                    ]
                    [
                        Limits of Layer-Based \\Latent CoT Interpretation
                        [
                            Proof of Turing completeness\\ in model architectures
                            [
                            RNN encoder-decoder~\citep{Cho2014LearningPhrase}{,} 
                            Transformers~\citep{vaswani2017attention}{,} 
                            Turing Completeness~\citep{perez2019turing}~\citep{qiu2024ask}
                            , leaf,text width=40em
                            ]
                        ]
                    ]
                ]
                [
                    Towards Infinite-depth\\ Reasoning (\S\ref{sec:infinite-depth reasoning})
                    [
                        Spatial Infinite Reasoning:\\ Text Diffusion Models
                        [
                            Masked Diffusion Models
                            [
                                D3PM~\citep{austin2021structured}{,} SEDD~\citep{lou2024discrete}{,} RADD~\citep{ou2024your}{,} MD4~\citep{shi2024simplified}{,} Simple-MDM~\citep{sahoo2024simple}{,} MDM~\citep{nie2024scaling}{,} \\
                                MMaDA~\citep{yang2025mmada}{,} IRED~\citep{Du_2024_ICML}{,} LLaDA~\citep{nie2025large}{,} dKV-Cache~\citep{ma2025dkv}{,} DoT-SEDD~\citep{ye2024diffusionthoughtschainofthoughtreasoning}{,} \\
                                dLLM-Cache~\citep{liu2025dllmcacheacceleratingdiffusionlarge}{,} d1-LLaDA~\citep{zhao2025d1scalingreasoningdiffusion}{,} MGDM~\citep{ye2025autoregressiondiscretediffusioncomplex}{,} LLaDA~1.5~\citep{zhu2025llada15variancereducedpreference}{,} DCoLT~\citep{huang2025reinforcingdiffusionchainlateral}
                                , leaf,text width=40em
                            ]
                        ]
                        [
                            Embedding-based Diffusion \\Models
                            [
                                Diffusion-LM~\citep{li2022diffusionlmimprovescontrollabletext}{,} CDCD~\citep{dieleman2022continuousdiffusioncategoricaldata}{,} Bit Diffusion~\citep{chen2023analogbitsgeneratingdiscrete}{,} Plaid~\citep{gulrajani2023likelihoodbaseddiffusionlanguagemodels}{,} BFN~\citep{graves2025bayesianflownetworks}{,} DoT-Plaid~\citep{ye2024diffusionthoughtschainofthoughtreasoning}{,} \\
                                TESS~\citep{mahabadi2024tesstexttotextselfconditionedsimplex}{,} TESS 2~\citep{tae2025tess}
                                , leaf,text width=40em
                            ]
                        ]
                        [
                            Hybrid AR-Diffusion Models
                            [
                                DiffuLLaMA~\citep{gong2025scaling}{,} Dream~\citep{dream2025}{,} L2D~\citep{cetin2025large}{,} Gemini Diffusion~\citep{google2025geminiDiffusion}{,} Mercury~\citep{labs2025mercuryultrafastlanguagemodels}
                                , leaf,text width=40em
                            ]
                        ]
                    ]
                    [
                        'Infinitely Long' \\Optimizer Network
                        [
                            Infini-attention
                            [
                            Transformers with infini-attention~\citep{munkhdalai2024leave}
                            , leaf,text width=40em
                            ]
                        ]
                        [
                            Test-time training (TTT)\\ and its descendants
                            [
                            TTT~\citep{sun2024ttt}{, }Titans~\citep{behrouz2024titans}{, }OmegaNet and Atlas~\citep{behrouz2025atlas}
                            , leaf,text width=40em
                            ]
                        ]
                        [
                            Implicit Fixed-Point RNNs
                            [
                            Implicit language models are rnns~\citep{schone2025implicit}
                            , leaf,text width=40em
                            ]
                        ]
                    ]
                ]
            ]
        \end{forest}
}
    \caption{Taxonomy of Latent Reasoning.}
    \label{fig:taxo_of_lr}
\end{figure}

\section{Preliminary: Latent Chain-of-Thought}
\label{sec:preliminary}

In this section, we present a unified mathematical framework for understanding various Latent CoT approaches. Unlike traditional CoT reasoning that generates explicit textual intermediate steps, latent CoT methods perform reasoning through continuous representations and hidden states within the model's computational graph. We categorize these approaches based on how they propagate information across layers (spatial dimension) and time steps (temporal dimension).


\subsection{General Framework}

We begin by establishing a general formulation for transformer-based reasoning systems. Consider a transformer model processing information at time step $t$ and layer $l$. Let $x_t^l \in \mathbb{R}^d$ denote the activation at layer $l$ and time $t$. We introduce $\mathbf{S}_t^l$ to represent the hidden state that captures historical information. The structure and dimensionality of $\mathbf{S}_t^l$ are architecture-dependent and define how context is maintained. This state can manifest in several forms, including:
\begin{itemize}
    \item \textbf{KV Cache:} In standard Transformers, $\mathbf{S}_t^l$ is the Key-Value (KV) cache, comprising a pair of matrices $(\mathbf{K}_t^l, \mathbf{V}_t^l)$, where $\mathbf{K}_l, \mathbf{V}_l \in \mathbb{R}^{n \times d}$ and $n$ is the sequence length of the context. Note that as $t$ increases, so does $n$.
    \item \textbf{Linear Attention State:} In models with linear attention, the hidden state can be compressed into a fixed-size state matrix, $\mathbf{S}_t^l \in \mathbb{R}^{d \times d}$, which allows for efficient, recurrent-style updates.
    \item \textbf{Recurrent State:} For RNN-like mechanisms, $\mathbf{S}_t^l$ is a single state vector, $\mathbf{S}_t^l \in \mathbb{R}^{d}$, that summarizes all past information into a fixed-size representation.
\end{itemize}

With this generalized view, the fundamental operations in latent reasoning can be decomposed into spatial and temporal transformations.

The spatial transformation propagates information vertically through layers at a fixed time step:
\begin{equation}
   \bm x_{t+1}^{l+1} = f(\bm x^l_{t+1}, g(\mathbf{S}^l_t, \bm x^l_t))
\end{equation}
where $f$ represents the layer-wise transformation function (e.g., a transformer block), which uses the historical context in $\mathbf{S}_t^l$ to compute the next layer's activation; $g$ captures how historical information is maintained and updated. The implementation of $g$ depends on the form of $\mathbf{S}_t^l$ (e.g., appending to the KV cache, or performing a matrix/vector update).

\paragraph{Activation-Based Methods} Activation-based methods focus on deepening the computational graph by iteratively refining activations within a single time step. These approaches implement a form of recursive computation where the same transformation is applied multiple times, allowing for progressive refinement of representations.

Formally, activation-based methods compute:
\begin{equation}
    \bm x_t^{l+n} = f\left( \dots f\left( f(\bm x_t^l, g(\mathbf{S}_t^l, \bm x_t^l)), g(\mathbf{S}_t^{l+1}, \bm x_t^{l+1}) \right), \dots, g(\mathbf{S}_t^{l+n-1}, \bm x_t^{l+n-1}) \right)
\end{equation}

This recursive application can be understood as creating a computational loop within the forward pass. At each iteration $i \in \{1, \ldots, n\}$, the model refines its representation by applying the transformation function $f$, potentially with access to different hidden states $\mathbf{S}_t^{l+i-1}$. Here, $l$ denotes the starting layer index, constrained by $1 \le l \le L-n$, where $L$ is the total number of layers in the model. The key insight is that by repeatedly processing the same input with shared parameters, the model can perform iterative refinement analogous to human step-by-step reasoning.

\paragraph{Hidden State-Based Methods} Hidden state-based methods take a fundamentally different approach by aggregating information from multiple temporal or spatial contexts simultaneously. Rather than iterative refinement, these methods leverage rich historical representations to inform current computations.

The core computation in hidden state-based methods is:
\begin{equation}
    \bm x_t^{l+1} = f\left(\bm x_t^l, g\left(\left(\mathbf{S}_t^l, \mathbf{S}_{t-1}^{l}, \ldots, \mathbf{S}_{t-n}^{l}\right), \bm x_t^l\right)\right),
\end{equation}

This operation allows the model to access a broader context of hidden states, effectively creating a memory bank that spans multiple layers or time steps. The function $f$ must be designed to effectively aggregate and utilize this expanded context, often through specialized attention mechanisms or learnable aggregation functions.

\subsection{Connections to Explicit Chain-of-Thought}

Understanding how these latent methods relate to explicit Chain-of-Thought reasoning provides important insights. Traditional CoT generates a sequence of tokens $y_1, y_2, \ldots, y_T$ representing intermediate reasoning steps. In the latent framework, these explicit tokens are replaced by continuous representations that evolve according to the dynamics described above.

The correspondence can be formalized by considering the generation process. In explicit CoT:
\begin{equation}
    \bm y_{t+1} = \text{Decode}(\text{Transform}(\bm x_t, \mathbf{S}_t)),
\end{equation}
where the decoding step projects continuous representations back to discrete tokens.

Latent methods eliminate this decoding step, instead maintaining reasoning in the continuous space:
\begin{equation}
    \bm z_{t+1} = \text{Transform}(\bm z_t, \mathbf{S}_t),
\end{equation}
where $z_t$ represents the continuous "thought" at step $t$.

This fundamental difference enables latent methods to explore reasoning pathways that may not have natural linguistic expressions, potentially discovering more efficient or powerful reasoning strategies unconstrained by the token vocabulary. However, it also introduces challenges in interpretability and training, as the intermediate states no longer correspond to human-readable explanations.

\subsection{Latent Reasoning Updates of Diffusion Models }

Understanding how latent update methods relate to diffusion models reveals fundamental differences from autoregressive (AR) generation. Traditional diffusion models operate purely through \textbf{temporal updates} without explicit spatial transformations, fundamentally differing from the spatial-temporal decomposition in transformer-based reasoning systems.

\paragraph{Temporal-Only Updates Diffusion Models}
Classical diffusion models perform updates exclusively in the temporal dimension through iterative denoising. The process involves two primary update mechanisms:

\textbf{Discrete updates (mask-based):} Given a sequence of tokens ${y_1,\dots,y_N}$, the model selectively updates positions based on masking patterns:
\begin{equation}
\bm x_{t+1}^l(i) = \begin{cases}
f(\bm x_t^l(i), \epsilon_t), & \text{if } m_t(i) = 1 \\
\bm x_t^l(i), & \text{otherwise}
\end{cases}
\end{equation}
where $m_t(i)$ represents the mask indicating which tokens to update at step $t$.

\textbf{Continuous updates (noise-based):} The model applies global noise reduction across all positions:
\begin{equation}
\bm x_{t+1}^l = f(\bm x_t^l, \epsilon_t)
\end{equation}
where $f$ represents the denoising function that operates uniformly across all token positions.

\paragraph{KV-cache Integrated Diffusion Models}
Recent advances have begun incorporating bidirectional KV cache mechanisms \cite{ma2025dkv} into diffusion models, introducing spatial-like transformations alongside temporal updates. This hybrid approach bridges the gap between traditional diffusion and transformer-based reasoning.

\textbf{Confidence-thresholded spatial transformation:}
All token activations are updated layer-wise at each denoising iteration:
\begin{equation}
\bm x_t^{l+1} = f_{\tau}\bigl(\bm x_t^l,\mathbf{S}t^l,\epsilon_t\bigr)
\end{equation}
where $f_{\tau}$ denotes a bidirectional Transformer block that refines every token representation while utilizing cached states.

\textbf{Selective temporal cache updates:}
Only tokens whose confidence score $c_t^l(i)=\mathrm{conf}\bigl(x_t^l(i)\bigr)$ meets or exceeds threshold $\tau$ refresh their KV cache:
\begin{equation}
\mathbf{S}_{t+1}^l(i) = \begin{cases}
g_{\tau}\bigl(x_t^l(i),\,\mathbf{S}_t^l(i)\bigr), & c_t^l(i)\ge\tau\\
\mathbf{S}_t^l(i), & \text{otherwise}
\end{cases}
\end{equation}

\textbf{Complete spatio-temporal evolution:}
The framework combines spatial refinement with selective temporal caching:
\begin{equation}
\bm x_{t+1}^{\,l+1} = f_{\tau}\bigl(\bm x_{t+1}^l,\,\mathbf{S}_{t+1}^l\bigr)
\end{equation}

This evolution represents a significant departure from traditional diffusion models, incorporating transformer-style spatial processing while maintaining the iterative refinement benefits of temporal diffusion. The confidence-thresholded mechanism enables efficient cache management in bidirectional contexts, addressing the fundamental incompatibility between traditional KV caching and diffusion model architectures.

\medskip

Consequently, diffusion models scan the entire sequence to identify and update the highest-confidence tokens in parallel---continuously correcting their representations across layers---whereas autoregressive models must commit to a single next token and cannot revisit or refine earlier outputs.
As a result, diffusion’s spatio–temporal mechanism enables ongoing, bidirectional refinement of multiple reliable latent states, while AR generation proceeds strictly forward, leaving past tokens fixed once generated.

\section{Latent Reasoning}
\label{sec:latent cot reasoning}

\begin{figure}[t]
    \centering
    \includegraphics[width=\linewidth]{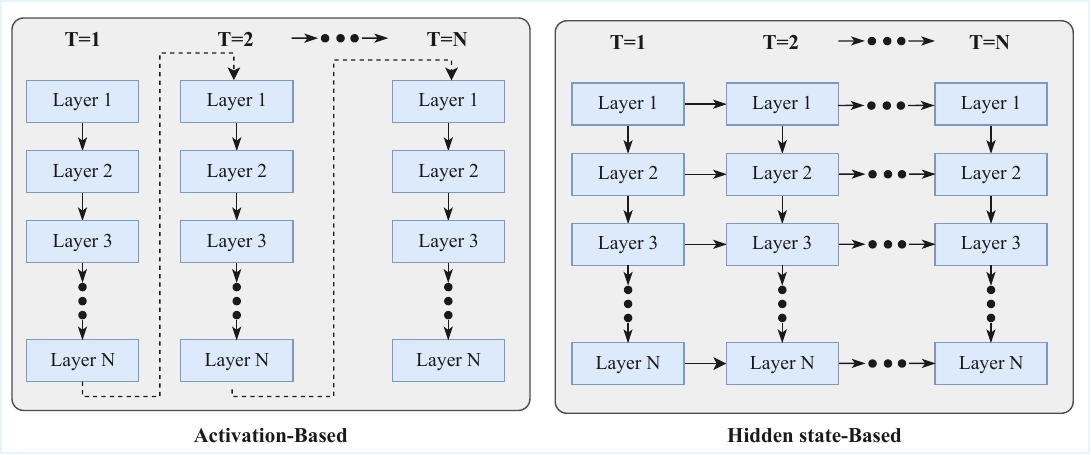}
    \caption{Comparison of Activation-Based and Hidden-state-Based Latent Reasoning. Activation-based methods (left) iteratively refine representations by looping through the same layers over multiple time steps ($T=1, 2,..., N$), increasing computational depth. Hidden-state-based methods (right) process information sequentially, evolving a hidden state that carries information across a potentially long temporal sequence ($T=1, 2,..., N$).}
    \label{fig:reasoning_paradigms}
\end{figure}

The development of latent CoT reasoning follows two fundamental computational paradigms: expanding depth through activation recurrence and expanding temporal capacity through hidden state evolution. As illustrated in Figure~\ref{fig:reasoning_paradigms}, activation-based methods create deeper computational graphs by iteratively processing information through the same set of layers, akin to vertical expansion. In contrast, hidden-state-based methods expand the model's memory horizontally, allowing it to access and integrate information over longer sequences.

This distinction raises critical implementation and theoretical questions. For activation-based approaches, \textbf{how can a model with a fixed number of layers be architecturally designed or trained to "think" longer about a problem, effectively creating vertical computational depth on the fly? What are the principles that govern this induced recurrence, and what new capabilities does it unlock?} Conversely, for hidden-state methods, as reasoning chains extend, \textbf{how can a model maintain a coherent "state of mind" over vast temporal sequences without succumbing to the bottleneck of ever-expanding memory? Can this temporal evolution be reframed as a form of continuous online optimization, conceptually unifying this horizontal expansion with the iterative vertical refinement seen in activation-based methods?}

While both approaches enhance reasoning capabilities, they differ in implementation requirements and deployment flexibility, offering distinct pathways toward more powerful latent reasoning. The following sections of this paper will describe these parts in detail.

\subsection{Vertical Recurrent: Activation-based Methods}

Activation-based approaches achieve latent reasoning by creating recurrent computational 
flows, either through architectural design or training-time manipulation. These methods 
share a common principle: iteratively refining representations without generating explicit 
reasoning tokens.

\subsubsection{Loop/Universal Transformer Recurrence}
\begin{table}[h]
  \raggedright
  \scriptsize
  \renewcommand{\arraystretch}{1.2}
  \setlength{\tabcolsep}{4pt}
  \resizebox{\textwidth}{13.5mm}{
  \begin{tabular}{lccccc}
    \toprule
    \textbf{Architecture} & \textbf{Pre/Loop/Coda} & \textbf{Per‑iter input $x_t$} & \textbf{Hidden state $S_t$} & \textbf{Dynamic stop} & \textbf{Depth‑emb $d_t$} \\
    \midrule
    Universal Transformer$^{\citep{dehghani2018universal}}$ & No & $x_t^{l-1}+d_t$ & standard unroll & ACT, $\sum\nolimits_t p_t>\tau$ & sinusoidal $d_t$ \\[0.35ex]

    CoTFormer$^{\citep{mohtashami2023cotformer}}$ & No & $S_t^{l-1}, x_t^{l-1}$ & standard unroll & MoR router $g_i$ & learnable $d_t$ \\[0.35ex]

    Recursive Transformer$^{\citep{bae2024relaxed}}$ & Optional & $x_t^{l-1}$ & share/refill $\hat h$ & early‑exit, $\max\nolimits_t\!\Delta h<\varepsilon$ & none \\[0.35ex]

    AlgoFormer$^{\citep{gao2024algoformer}}$ & Yes & $x_t^{l-1}$ & standard roll & fixed & none \\[0.35ex]

    Recurrent‑Depth$^{\citep{geiping2025scaling}}$ & Yes & $x_t^{1},x_t^{l-1}$ & modulo $(t\bmod r)$ reuse & fixed‑point iteration & tried, dropped \\
    \bottomrule
  \end{tabular}
  }
  \caption{Comparison of activation-based latent CoT architectures and their key design characteristics, showing the evolution from early monolithic designs to structured Pre/Loop/Coda frameworks with simplified dynamic stopping mechanisms.}
  \label{tab:activation_based_comparison}
\end{table}

\begin{figure}[h!]
  \centering
  \small
  \definecolor{pre_color}{RGB}{220,223,240}
  \definecolor{loop_color}{RGB}{203,231,207}
  \definecolor{coda_color}{RGB}{252,226,187}
  \definecolor{op_color}{RGB}{255,220,220}
  \definecolor{hid_color}{RGB}{194,232,247}
  \resizebox{0.95\textwidth}{!}{
  \begin{tikzpicture}[
      stage/.style={
        draw=black,
        very thick,
        minimum width=3.2cm,
        minimum height=1.5cm,
        rounded corners=8pt,
        font=\large
      },
      plus/.style={
        draw=black,
        very thick,
        circle,
        fill=op_color,
        minimum size=12pt,
        inner sep=0pt,
        outer sep=0pt,
        font=\small
      },
      gate/.style={
        draw=black,
        very thick,
        circle,
        fill=op_color,
        minimum size=12pt,
        inner sep=0pt,
        outer sep=0pt,
        font=\footnotesize
      },
      link/.style={-Latex,very thick,rounded corners=4pt},
      dashedlink/.style={-Latex,very thick,densely dashed,rounded corners=4pt},
      mem/.style={
        draw=black,
        very thick,
        fill=hid_color,
        rounded corners=6pt,
        minimum height=1cm,
        minimum width=2cm,
        font=\large
      }
    ]
    
    \node[stage, fill=pre_color] (pre) {Prelude};
    \node[stage, fill=loop_color, right=3cm of pre] (loop) {Loop blocks};
    \node[stage, fill=coda_color, right=3cm of loop] (coda) {Coda};
    
    \node[plus, left=2cm of pre] (add) {$+$};
    \node[above=1cm of add] (input) {Input tokens};
    \node[below=0.8cm of add] (dt) {$d_t$};
    
    \draw[link] (input) -- (add);
    \draw[link] (dt) -- (add);
    \draw[link] (add) -- (pre);
    
    \draw[link] (pre) -- (loop);
    \draw[link] (loop) -- (coda);
    
    \node[right=2cm of coda] (output) {Output};
    \draw[link] (coda) -- (output);
    
    \node[mem, below=2.5cm of loop] (hmem) {$S_t^{1},S_t^{2},...,S_t^{l}$};
    
    \draw[link] (loop.south) -- ++(0,-0.8) -| (hmem.north);
    \draw[link] (hmem.north) -- ++(0,0.8) -| (loop.south);
    
    \node[right=0.5cm of hmem, font=\large] (kvcache) {KV-cache};
    \draw[dashedlink] (hmem.east) -- (kvcache.west);
    
    \node[gate, right=0.8cm of loop] (stopgate) {$\sigma$};
    \node[below=0.2cm of stopgate, font=\footnotesize] {stop};
    \draw[link] (loop.east) -- (stopgate.west);
    
    \node[above=0.4cm of loop, font=\large\bfseries] {$l = 1, 2, \ldots, N$};
    \node[above=0.1cm of loop, font=\small] {iterations};
    
    \node[
      draw=black,
      rounded corners=4pt,
      align=left,
      font=\small,
      anchor=south east,
      fill=white,
      fill opacity=0.9,
      text opacity=1,
      inner sep=8pt
    ] at ($(current bounding box.south east)+(-0.3,0.3)$) {%
      \textbf{Legend:}\\[2pt]
      $S_t$: hidden state / KV\\
      $d_t$: depth embedding\\
      $\sigma$: dynamic stop gate
    };
    
  \end{tikzpicture}
  }
  \caption{\small Conceptual diagram of a Pre/Loop/Coda architecture with per-iteration input $x_t$, hidden state $S_t$ (KV-cache), depth embedding $d_t$, and a dynamic-stop gate.}
  \label{fig:pre_loop_coda_diagram}
\end{figure}
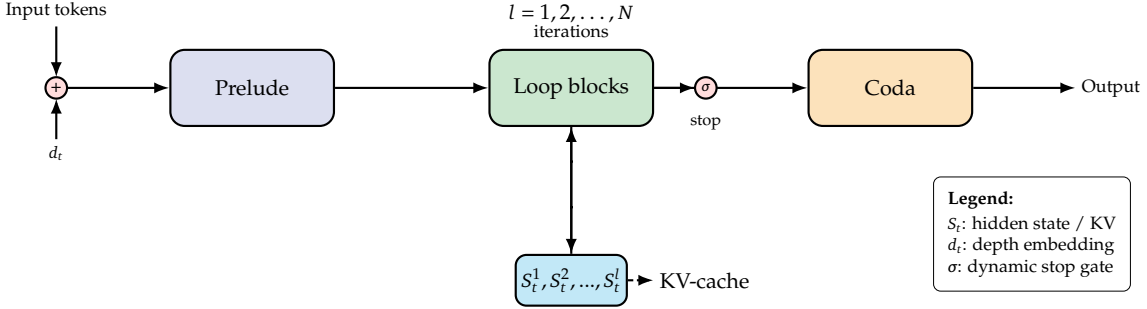

Loop-based architectures represent the foundational approach to activation-based latent CoT reasoning, implementing continuous activation propagation across Transformer layers through explicit architectural modifications. These models share a core principle: enabling iterative refinement of hidden states within a single forward pass through layer-wise recurrence. Starting from the Universal Transformer (UT)~\citep{dehghani2018universal}, which pioneered dynamic recurrence over layers with its Adaptive Computation Time (ACT) mechanism, this architectural paradigm has established depth-adaptive reasoning as a viable alternative to traditional fixed-depth transformers. The key innovation lies in treating network depth not as a static hyperparameter but as a dynamic computational resource that can be allocated based on task complexity. Extending activation-reuse beyond Universal/Looped Transformers,~\citet{zeng2025pretraining} introduce a Pondering LM that performs $k$ iterative ‘ponder’ cycles inside every token prediction. Each cycle converts the model’s softmax into a continuous pondering embedding: a weighted sum of all vocabulary vectors, which is fed back via a residual path to refine the hidden state.

Since this seminal work, the field has undergone systematic evolution along several key dimensions, revealing important design principles for latent reasoning architectures (Table~\ref{tab:activation_based_comparison} and Figure~\ref{fig:pre_loop_coda_diagram}).

\paragraph{The Rise of Pre/Loop/Coda Structure} Early models like Universal Transformer and CoTFormer~\citep{mohtashami2023cotformer} adopted monolithic recurrent designs without explicit stage separation. However, recent architectures like Recursive Transformer~\citep{bae2024relaxed}, AlgoFormer~\citep{gao2024algoformer}, and Recurrent-Depth~\citep{geiping2025scaling} have converged on a three-stage Pre/Loop/Coda structure. This design explicitly separates input encoding (Prelude), iterative reasoning (Loop blocks), and output decoding (Coda), enabling more modular and interpretable computation flows. The modularization of the architecture improves interpretability and facilitates the injection of task-specific priors, such as fixed-point iteration constraints or algorithmic templates, into the reasoning process.

\paragraph{Per-iteration Input and Hidden State Management} Input handling strategies vary across models, reflecting different hypotheses about information flow during recurrence. Universal Transformer combines previous layer output $x_t^{l-1}$ with depth embedding $d_t$. CoTFormer uses both hidden state $\mathbf{S}_t^{l-1}$ and $x_t^{l-1}$, while Recursive Transformer and AlgoFormer simplify to just $x_t^{l-1}$. Recurrent-Depth adopts a hybrid approach with both $x_t^{1}$ and $x_t^{l-1}$.

For hidden state management, most models use standard unrolling of KV caches. Notable exceptions include Recursive Transformer's share/refill mechanism and Recurrent-Depth's modulo-based reuse $(t\bmod r)$, which improve memory efficiency through periodic cache recycling, as shown in Table~\ref{tab:activation_based_comparison}. These innovations strike a balance between preserving temporal coherence and managing computational resources.

\paragraph{The Decline of Depth Embeddings} Depth embeddings show a clear deprecation trend. Universal Transformer introduced sinusoidal $d_t$, and CoTFormer experimented with learnable embeddings. However, subsequent models like Recursive Transformer and AlgoFormer completely dropped them. Recurrent-Depth tried but ultimately abandoned depth embeddings, suggesting their limited utility in recurrent architectures despite initial enthusiasm. This trend indicates that explicit positional encoding of depth may be redundant when the architecture inherently encodes iteration count through state evolution.

\paragraph{Simplification of Dynamic Stopping Mechanisms} Dynamic stopping mechanisms exhibit a clear trend toward simplicity. Universal Transformer's sophisticated ACT mechanism (with cumulative probability $\sum_t p_t>\tau$) gave way to CoTFormer's MoR router $g_i$. Recent models adopt even simpler strategies: Recursive Transformer uses early-exit based on change magnitude ($\max_t\Delta h<\varepsilon$), AlgoFormer opts for fixed iterations, and Recurrent-Depth explores fixed-point criteria. This evolution suggests that complex adaptive mechanisms may not justify their computational overhead in practice.

These architectural trends reflect the field's maturing understanding: moving from complex adaptive mechanisms toward stable, modular designs while preserving the core benefit of enhanced reasoning through layer-wise recurrence. The convergence on simpler, more interpretable designs suggests that the key to latent reasoning may lie not in sophisticated control mechanisms but in providing sufficient computational depth with efficient resource management.

\subsubsection{Activation with Explicit Hidden-State Feedback}
\label{subsec:act_hidden_feedback}

\noindent
While loop-based architectures refine token representations by rerunning the same set of layers, a distinct family of models \emph{feeds hidden states back into the input stream} between iterations. In these systems the hidden activations themselves become new sequence elements, so each recurrent step simultaneously extends the effective depth \textbf{and} exposes internal computation to subsequent attention.

\paragraph{Coconut} Proposed by \citet{hao2024training}, \textbf{Coconut} inserts a \emph{continuous thought} vector—the last-layer hidden state of the previous decoding step—as an extra position before the current token. Pondering therefore occurs in latent space without emitting textual reasoning, enabling breadth‑first exploration while reusing the same Transformer parameters.

\paragraph{CoTFormer} In \textbf{CoTFormer}~\citep{mohtashami2023cotformer}, every forward pass first computes preliminary token embeddings; these activations are then \emph{interleaved} back into the sequence and the shared block stack is executed again. Early‑exited tokens thus attend to deeper refinements of their own representations, realizing adaptive depth with minimal parameters.

Both models share three properties that distinguish them from “pure” activation‑based recurrence:

\paragraph{Key characteristics.} Explicit state tokens re-inject hidden vectors as sequence elements, bridging vertical recurrence and horizontal memory; no architectural expansion—the model reuses the same layers so parameter count stays constant while depth grows dynamically; and latent reasoning remains internal, thereby avoiding the latency of producing explicit CoT tokens.

These designs demonstrate that passing hidden states across recurrent hops can unlock stronger reasoning while preserving the efficiency of shared‑weight loops, and they foreshadow later hybrids that blend activation and hidden‑state paradigms.

\subsubsection{Training-induced Recurrence}

While architectural recurrence requires explicit structural modifications, an alternative pathway achieves similar computational benefits through specialized training on standard transformer architectures. These methods fundamentally create recurrent activation flows without changing the model's underlying structure, demonstrating that the key insight of iterative refinement can be induced through training alone. This approach is particularly valuable as it enables existing pretrained models to develop latent reasoning capabilities without architectural constraints.

The core principle unifying these methods is the creation of implicit loops in the computation graph: whether by feeding activations back into the model (continuous recurrence), compressing multi-step reasoning into iteratively-processed representations (compressed states), or extending the effective computation depth through strategic token insertion (expanded iterations). All these approaches share the goal of enabling deeper reasoning without explicit architectural loops.

\paragraph{Continuous Activation Recurrence}

The most direct form of training-induced recurrence involves creating explicit loops of continuous activations. Ref.~\citep{hao2024training} pioneers this approach with Coconut, which loops the LLM's last hidden state (the "continuous thought") directly back into the model as input for the next step. This mechanism creates a recurrence pattern strikingly similar to architectural approaches like Universal Transformer, but implemented entirely through training. The continuous thought can encode multiple reasoning paths simultaneously, enabling breadth-first search-like exploration in latent space.

Building on this foundation, subsequent work has refined the training methodology while maintaining the core recurrence principle. \citet{shen2025codi} propose CODI, which frames the problem as learning to align recurrent hidden states through self-distillation. By aligning the hidden activation before the final answer between teacher (with full CoT) and student (with compressed reasoning) paths, CODI effectively learns a fixed-point iteration in activation space. This single-step alignment proves more stable than Coconut's curriculum learning, achieving parity with explicit CoT on GSM8K for the first time among latent methods.

\citet{cheng2024compressed} take a different approach with CCOT, training the model to generate variable-length sequences of continuous embeddings that approximate full reasoning traces. These embeddings function as compressed representations of recurrent computation steps, maintaining the iterative nature while reducing sequence length. The optional decoding back to text preserves interpretability while confirming that meaningful computation occurs in these latent iterations. PCCOT~\citep{wu2025parallel} uses Jacobi-iteration allowing parallel continuous thoughts. Building on pause- and filler-token methods that prolong hidden-state computation, System-1.5 Reasoning~\cite{wang2025system1.5} introduces Depth and Step Shortcuts that dynamically allocate vertical layer depth and horizontal reasoning steps, delivering over $20\times$ faster inference on GSM8K while preserving chain-of-thought accuracy—all without modifying the Transformer backbone.

\paragraph{Compressed State Recurrence}

Rather than continuous loops, another strategy compresses reasoning steps into discrete or semi-discrete representations that the model processes recurrently. \citet{su2025token} replace early CoT segments with discrete latent tokens learned via VQ-VAE, creating "assorted" reasoning that mixes compressed abstract steps with detailed reasoning. This approach effectively creates a hierarchical recurrence where abstract tokens trigger expanded computation in subsequent layers.

\citet{zhang2025lightthinker} employ "gist tokens" as compression anchors in hidden space. Though these tokens themselves are semantically meaningless, they serve as recurrence checkpoints where the model aggregates and redistributes computational state. The attention mask manipulation enforces that subsequent reasoning depends on these compressed states, creating an implicit recurrence structure through the sequence.

The key insight across these compression methods is that they transform horizontal (sequence-level) reasoning into vertical (depth-level) computation, effectively increasing the recurrence depth available for each logical step.

\paragraph{Iteration Expansion through Strategic Tokens}

A third category of training-induced recurrence works by expanding the number of implicit iterations through token insertion. This approach recognizes that additional tokens, even without explicit semantic content, can provide more recurrence steps for internal computation.

\citet{pfau2024lets} demonstrate that even meaningless filler tokens (e.g., "......") can improve reasoning by simply providing more attention steps, effectively increasing the number of recurrent iterations the model can perform. ~\citet{goyal2024think} refine this with learnable `<pause>` tokens that explicitly signal computation steps, creating trainable recurrence points that the model learns to utilize effectively.

More sophisticated approaches inject structured tokens that organize the recurrence pattern. \citet{wang2024guiding} introduce planning tokens that create a hierarchical recurrence structure, where each planning token initiates a new reasoning loop with specific computational goals. \citet{jin2024disentangling} further decompose reasoning into `<memory>` and `<reason>` tokens, creating specialized recurrence patterns for different types of cognitive operations. These structured approaches demonstrate that training can induce not just recurrence, but organized, interpretable recurrence patterns.

\paragraph{Implications and Connections} 

These training-induced methods reveal a fundamental insight: recurrence for reasoning is not solely an architectural property but can emerge from appropriate training objectives. The success of these approaches suggests that standard transformers possess latent capacity for iterative computation that training can unlock. Moreover, the convergence of continuous, compressed, and token-based methods toward similar performance outcomes indicates that the specific implementation of recurrence matters less than ensuring sufficient computational depth for reasoning tasks.

The relationship between these training-induced methods and architectural recurrence is complementary rather than competitive. Future work might explore hybrid approaches that combine architectural loops with training-induced recurrence patterns, potentially achieving the benefits of both explicit structure and learned optimization.

\subsubsection{Training Strategies for Recurrent Reasoning}

Effectively training models with recurrent activation flows presents unique challenges, as these architectures must learn to leverage iterative computation rather than relying solely on feedforward depth. Researchers have developed specialized training strategies that address both architectural and induced recurrence.

For architectural recurrence, MIDAS~\citep{saunshi2024inductive} proposes a progressive stacking framework to address training stability in loop-based models. It defines a replication operator $\mathcal{M}(f, b)$ that duplicates the middle layers of a base model $f$ by a factor $b$, enabling gradual depth expansion. Training proceeds through $k$ stages where model depth increases progressively, with each deeper model initialized from the previous stage. This curriculum approach helps models develop stable iterative reasoning patterns. Complementing this architectural focus, \citet{saunshi2025reasoning} introduce a looping-inspired regularization that enables even standard Transformers to benefit from recurrence-like properties through a cosine-similarity term $\mathcal{R}_G(k)$ in the loss function. This approach reveals that recurrent behavior can emerge from appropriate training objectives alone.

For training-induced recurrence, Stepwise Internalization~\citep{deng2024explicit} pioneered curriculum-based compression of reasoning traces. This technique gradually removes CoT tokens during fine-tuning, allowing models to internalize reasoning patterns into their parameters. This curriculum principle has been widely adopted, notably by Coconut~\citep{hao2024training} which progressively replaces CoT tokens with continuous thoughts, achieving fully latent inference loops. RELAY~\citep{yu2025enhancing} takes a more direct approach by explicitly aligning recurrence steps with reasoning steps through a two-stage process: first training looped Transformers with CoT-aligned supervision using loss $\mathcal{L}=\mathcal{L}_{ans}+\lambda \mathcal{L}_{iter}$, then fine-tuning autoregressive models on the generated reasoning chains.

These diverse training strategies converge on key principles: gradual complexity increase, alignment between recurrence depth and reasoning steps, and careful balance between architectural constraints and learned behaviors. The success of both architectural and training-induced approaches suggests that effective recurrent reasoning emerges from the interplay of structure and optimization.

\subsubsection{Applications and Capabilities}

The true test of recurrent reasoning methods lies in their ability to tackle complex tasks requiring structured, multi-step computation. Both architectural and training-induced recurrence have demonstrated remarkable capabilities across diverse domains.

In algorithmic generalization, recurrent models exhibit unprecedented extrapolation abilities. \citet{schwarzschild2021can} and \citet{giannou2023looped} demonstrate that looped architectures can generalize from small problem instances to significantly harder ones by extending recurrence steps at test time—a property unavailable to static-depth Transformers. This recurrence-controlled scaling mimics human-like progressive problem-solving and has been formalized through theoretical frameworks of looped computation graphs. Similarly, training-induced methods like Coconut show that continuous thought loops can solve logical reasoning tasks (ProsQA, PrOntoQA) through latent breadth-first search, while compressed-state methods achieve parity with explicit CoT on mathematical reasoning (GSM8K).

In symbolic reasoning and graph algorithms, recurrent models bridge neural and algorithmic computation. \citet{de2024simulation} show that looped Transformers with graph-specific attention heads can simulate classical algorithms (BFS, DFS, shortest-path) within bounded memory. This capability extends to training-induced recurrence: models with planning tokens~\cite{wang2024guiding} demonstrate improved performance on multi-hop reasoning by creating hierarchical computation structures. The decomposition of reasoning into specialized tokens (<memory>, <reason>)~\citep{jin2024disentangling} further enhances performance on tasks requiring both retrieval and logical inference.

In optimization and meta-learning, works like~\cite{gatmiry2024can, chen2024bypassing} prove that looped models implicitly implement multi-step gradient descent, revealing deep connections between recurrence and optimization. This theoretical insight explains why both architectural loops and training-induced continuous thoughts converge on similar computational patterns: they are fundamentally performing iterative refinement analogous to optimization algorithms.

These applications demonstrate that recurrent reasoning—whether achieved through architecture or training—provides a general framework for complex computation. The convergence of different approaches on similar capabilities suggests that the key insight is not the specific implementation but ensuring sufficient iterative depth for the task at hand.

\subsection{Horizontal Recurrent: Hidden state-based Methods}

As previously mentioned, activation-based approaches focus on expanding layer depth in networks. However, deeper networks inevitably encounter challenges such as gradient explosion or vanishing. In contrast, the temporal dimension can be readily expanded to millions of tokens. From a theoretical perspective, the temporal dimension can also be conceptualized as a form of depth, which raises an important research question: \textbf{How can we effectively expand the latent reasoning process along the temporal dimension?}

The standard Transformer provides a baseline for this horizontal expansion. It handles temporal information by storing all previous token inputs as key-value pairs in what is known as the 
KV cache. This cache effectively serves as the model's hidden state, preserving a rich history of the sequence. However, this approach has a critical bottleneck: the KV cache grows linearly with the sequence length, leading to unbounded memory consumption that makes processing very long sequences impractical.

 To address this challenge, we can compress previous information into a fixed-size vector or matrix, similar to RNNs. When working with hidden states, there are two primary approaches to enhance their expressiveness: (1) the Linear-State recurrence approach, which applies update and decay rules to the hidden states, and (2) Gradient-State recurrence approach, treating hidden states as online-learning parameters and optimizing them using online learning methods. \textbf{Notably, although these methods have not yet produced evidence demonstrating enhanced reasoning capabilities, their theoretical properties suggest they may play a significant role in the future, as they represent a form of iterative processing that is conceptually similar to layer stacking.}

\subsubsection{Linear-State Recurrence}

For the first approach, models such as Mamba-2~\citep{dao2024transformers}, GLA~\citep{yang2023gated}, RWKV-6~\citep{peng2024eagle}, and HGRN2~\citep{qin2024hgrn2} represent early attempts in this direction. 
A matrix-valued hidden state $S$ is transmitted and updated along the temporal dimension. At each time step, the hidden state undergoes global decay, followed by updates incorporating information from the current time step. 

Remarkably, these diverse linear attention models can be unified under a general framework of associative recurrent neural networks with matrix-valued hidden states~\citep{yang2024deltanet,yang2024gateddeltanet}. Given a matrix-valued hidden state $\mathbf{S}_t \in \mathbb{R}^{d \times n}$ and current input $\mathbf{x}_t \in \mathbb{R}^{d}$, these models follow the general form:
\begin{align}
\mathbf{S}_t &= \mathbf{S}_{t-1} + \bm {k_t v_t}^\top, \quad \text{(recurrence)} \\
\mathbf{o}_t &= \mathbf{S}_t \bm{q}_t, \quad \text{(memory read-out)}
\end{align}
where $\bullet$ represents an associative operator (e.g., Hadamard product, matrix multiplication), and $\mathbf{M}_t$, $\bm {v}_t$, $\bm {k}_t$, $\bm{q}_t$ are functions of the current input $\bm{x}_t$. The use of associative operators enables parallel scan calculations of $\mathbf{S}_1, \dots, \mathbf{S}_L$, facilitating efficient training. Table~\ref{tab:unified_memory_update} illustrates how various models instantiate this framework.

However, a more profound perspective emerges when interpreting this state evolution through the lens of online optimization gradient. A key insight comes from DeltaNet~\citep{yang2024deltanet}, which perfectly exemplifies this duality. While its state update rule has a closed-form algebraic expression (see Table~\ref{tab:unified_memory_update} linear recurrent attention part), it is mathematically equivalent to applying a single gradient descent step to an online regression objective $\mathcal{L}(\bm S) = \frac{1}{2}\|\bm S\bm k_t-\bm v_t\|_2^{2}$. 

This gradient-state recurrence view is conceptually transformative. It reframes the temporal evolution of the hidden state $\bm S_t$ as a form of iterative refinement, akin to training a neural network layer. In this sense, the state matrix $\bm S$ is effectively treated as a dynamic, "fast weight" layer that is updated at each step based on a local objective. This perspective conceptually unifies the "temporal" recurrence of hidden-state models with the "depth" recurrence of activation-based models, suggesting a shared underlying principle of iterative processing for latent reasoning.

\subsubsection{Gradient-State Recurrence} 

\label{sec:opt-state-evo}

\begin{table}[htbp]
  \centering
  \small
  \begin{threeparttable}
  \begin{tabular}{ll}
    \toprule
    \textbf{Method} & \textbf{Unified Memory-update Rule} \\
    \midrule
    \multicolumn{2}{l}{\emph{Linear-State Recurrence}} \\
    Linear Attn~\citep{katharopoulos2020transformers} &
      $\bm S_t = \bm S_{t-1} + \bm k_t \bm v_t^{\!\top}$ \\

    RetNet/Lightning~\citep{sun2023retentive} &
      $\bm S_t = \gamma \bm S_{t-1} + \bm k_t \bm v_t^{\!\top}$ \\

    GLA~\citep{yang2023gated} &
      $\bm S_t = \bm S_{t-1}\!\operatorname{Diag}(\bm a_t) + \bm k_t \bm v_t^{\!\top}$ \\



    Mamba-2~\citep{dao2024transformers} &
      $\bm S_t = \alpha_t\,\bm S_{t-1} + b_t\,\bm k_t \bm v_t^{\!\top}$ \\

    HGRN-2~\citep{qin2024hgrn2} &
      $\bm S_t = \bm S_{t-1}\!\operatorname{Diag}(\bm a_t)
                +(\bm 1-\bm a_t)\bm v_t^{\!\top}$ \\

\midrule
\multicolumn{2}{l}{\emph{Linear/Gradient-State duality}} \\

DeltaNet~\citep{yang2024deltanet} &
\begin{tabular}[t]{@{}l@{}}
\textit{State-update:} $\displaystyle
  \bm S_t=\bm S_{t-1}(I-\beta_t\bm k_t\bm k_t^{\!\top})
          +\beta_t\bm k_t\bm v_t^{\!\top}$ \\[2pt]
\textit{Optimization:} $\displaystyle
  \bm S_t=\bm S_{t-1}-\beta_t\nabla_{\!\bm S}
  \tfrac12\|\bm S\bm k_t-\bm v_t\|_2^{2}$
\end{tabular} \\

G-DeltaNet~\citep{yang2024gateddeltanet} &
\begin{tabular}[t]{@{}l@{}}
\textit{State-update:} $\displaystyle
  \bm S_t=\alpha_t\bm S_{t-1}(I-\beta_t\bm k_t\bm k_t^{\!\top})
          +\beta_t\bm k_t\bm v_t^{\!\top}$ \\[2pt]
\textit{Optimization:} $\displaystyle
  \bm S_t=\bm S_{t-1}-\beta_t\nabla_{\!\bm S}
  \tfrac12\|\bm S\bm k_t-\bm v_t\|_2^{2}
  +\lambda\|\bm S-\alpha_t\bm S_{t-1}\|_F^{2}$
\end{tabular} \\[2pt]
    \midrule
    \multicolumn{2}{l}{\emph{Gradient-State Recurrence}} \\

    TTT~\citep{sun2024ttt} &
      $\bm S_t = \bm S_{t-1}
                 - \eta_t\,\nabla_{\!\bm S}\ell(\bm S_{t-1};\bm k_t,\bm v_t)$ \\

    Titans~\citep{behrouz2024titans}\tnote{*} &
      $\bm S_t = \alpha_t\,\bm S_{t-1}
                 - \eta_t\,\nabla_{\!\bm S}\ell(\bm S_{t-1};\bm k_t,\bm v_t)$ \\

    Lattice (orth.) ~\citep{karami2025lattice} &
      $\bm S_{i,t}=\bm S_{i,t-1}+\alpha_{i,t}\Bigl(I-
        \tfrac{\bm S_{i,t-1}\bm S_{i,t-1}^{\!\top}}
              {\lVert\bm S_{i,t-1}\rVert^{2}}\Bigr)\bm h_t$ \\

    Moneta~\citep{behrouz2025allc} &
      $\bm S_t=\operatorname{Norm}_q\!\bigl(
        \alpha_t\bm S_{t-1}-\eta_t\nabla_{\!\bm S}\ell_p(\bm S_{t-1};\bm k_t,\bm v_t)\bigr)$ \\

    Yaad (Huber)~\citep{behrouz2025allc} &
      $\bm S_t=a_t\bm S_{t-1}-\eta_t
        \begin{cases}
          \nabla_{\!\bm S}\ell_2,&\lVert S(\bm k_t)-\bm v_t\rVert\le\delta_t\\
          \delta_t\,\nabla_{\!\bm S}\ell_1,&\text{otherwise}
        \end{cases}$ \\

    Memora~\citep{behrouz2025allc} &
      $\bm S_t=\operatorname{Softmax}\!\bigl(
        \alpha_t\log\bm S_{t-1}-\eta_t\nabla_{\!\bm S}\ell_2(\bm S_{t-1};\bm k_t,\bm v_t)\bigr)$ \\

    OmegaNet~\citep{behrouz2025atlas} &
      $\displaystyle
        \bm S_t=\alpha_t\bm S_{t-1}-\!\!\sum_{i=t-c+1}^{t}\!\gamma_i
        \nabla_{\!\bm S}\bigl\|
        \bm S_{t-1}\phi(\bm k_i)-\bm v_i\bigr\|_2^{2}$ \\

    Atlas~\citep{behrouz2025atlas} &
      $\displaystyle
        \begin{aligned}
          \bm S^{\mathrm{aux}}_t &=
            \theta_t\bm S^{\mathrm{aux}}_{t-1}-\!\!\sum_{i=t-c+1}^{t}\!\eta_i
            \nabla_{\!\bm S}\bigl\|
            \bm S_{t-1}\phi(\bm k_i)-\bm v_i\bigr\|_2^{2}\\
          \bm S_t &=\alpha_t\bm S_{t-1}+\operatorname{NS5}(\bm S^{\mathrm{aux}}_t)
        \end{aligned}$ \\[4pt]

    \bottomrule
  \end{tabular}

  \begin{tablenotes}
    \item[*]\footnotesize Titans omits momentum and norm-adaptation terms for brevity.
  \end{tablenotes}

 \caption{\textbf{Unified hidden-state and optimization-based memory updates.}
    Each model is a recurrence on matrix memory $\bm S_t$: apply decay,
    projection or an optimization step to $\bm S_{t-1}$, then add an
    outer-product or gradient correction (read-out:
    $\bm o_t=\bm S_t\bm q_t$).
    \emph{Symbols:}
    For uniformity, $\alpha_t$ generally denotes the gate controlling the retention of the previous state, while $\eta_t$ denotes the learning rate. An important exception is \textbf{DeltaNet} and \textbf{Gated-DeltaNet}, whose learning rate or writing strength is denoted by~$\beta_t$.
    Additionally, $\gamma_i$ is the weight for each token's gradient in the \textbf{OmegaNet} context window, and $\theta_t$ is the momentum decay term in \textbf{Atlas}.
    All these parameters are data-/channel-dependent scalars, typically in $(0,1)$.
    $\delta_t$ is the Huber threshold;
    $\nabla\ell_p,\nabla\ell_1,\nabla\ell_2$ are gradients w.r.t.\
    $(\bm k_t,\bm v_t)$;
    $\operatorname{Norm}_q(\cdot)$ is $q$-norm normalization (Moneta);
    $\phi(\cdot)$ denotes a polynomial/high-order feature map;
    $\operatorname{NS5}(\cdot)$ is the Muon/Newton–Schulz 2\textsuperscript{nd}-order
    update (Atlas);
    $I-\frac{\bm s\bm s^{\!\top}}{\lVert\bm s\rVert^{2}}$ is the orthogonal
    projector in Lattice.}
    \label{tab:unified_memory_update}
  \end{threeparttable}
\end{table}

While linear-state models rely on predetermined decay–add rules, \emph{gradient-state} methods treat the hidden matrix as a set of fast-adapting parameters updated by a learnable optimizer. Each token triggers a lightweight descent step that steers the state toward the current key–value target, allowing the model to internalize task-specific dynamics on the fly. This view shifts the design space from choosing fixed linear kernels to selecting optimization algorithms (SGD, Adam-like, second-order, etc.), opening a rich continuum of memory behaviors governed by learning-rate schedules, momentum terms and higher-order corrections.

This insight paved the way for a second research trajectory that abandons closed-form descriptions entirely, in favor of direct online learning formulations~\citep{behrouz2025allc, behrouz2024titans, sun2024ttt, karami2025lattice, behrouz2025atlas}. This line of work, progressing from TTT (implementing SGD-like dynamics)~\citep{sun2024ttt} to Titans (incorporating Adam-like behaviors)~\citep{behrouz2024titans} and ATLAS (utilizing Muon optimization principles)~\citep{behrouz2025atlas}, formulates the state update explicitly as a gradient-based optimization step. Extending this optimization perspective, Ref.~\cite{li2025seek} introduce LATENTSEEK, a framework that performs test-time instance-level adaptation by directly optimizing latent representations using policy gradient.Despite their different origins, these approaches converge conceptually and can be understood through the general update rule:
\begin{equation}
\mathbf{S}_t = \alpha_t\mathbf{S}_{t-1} - \eta_t\nabla_{\!\bm S}\ell(\bm S_{t-1};\bm k_t,\bm v_t) 
\end{equation}
While powerful, this approach introduces significant challenges for parallelization. Unlike linear recurrent models that can be parallelized efficiently with a single scan operation, the gradient $\nabla\ell$ at step $t$ depends on the previous state $\bm S_{t-1}$. This inherent sequential dependency prevents parallel computation across the entire sequence length. Furthermore, these recurrent updates are embedded within complex architectural blocks that include standard components like LayerNorm and residual connections, making it difficult to fuse the computation into a single, hardware-efficient kernel.

To overcome these limitations, a practical solution known as \textbf{chunk-wise parallelization} has been widely adopted~\citep{behrouz2024titans, sun2024ttt, yang2024gateddeltanet}. This strategy balances expressiveness and efficiency:
\begin{itemize}
    \item \textbf{Intra-chunk Parallelism}: Within a small, fixed-size block (chunk) of the sequence, the gradients for all tokens are computed in parallel with respect to the \emph{same initial state} (the final state of the previous chunk). This breaks the sequential dependency within the chunk, allowing for efficient, batched computation.
    \item \textbf{Inter-chunk Recurrence}: The overall sequential nature of the model is maintained between chunks. The final state of one chunk is passed recurrently to become the initial state for the next, forming a chain at the chunk level.
\end{itemize}
Extending the optimization perspective beyond internal state updates, ~\citet{zhu2025soft} introduce Soft Reasoning, which treats the first-token embedding as a controllable latent variable. By injecting Gaussian noise and maximizing an Expected-Improvement objective via Bayesian optimization, the method dynamically searches the hidden space for a reasoning trajectory.

Although current research has not yet produced evidence demonstrating enhanced reasoning capabilities in these models, their theoretical properties suggest significant potential, particularly for enabling self-iteration in the absence of input tokens.

\subsubsection{Training-induced Hidden-State Conversion}

Building on the success of \emph{training‑induced recurrence} for activation‑based models, a parallel line of work shows that \
\textbf{fixed‑architecture Transformers can be \emph{converted}, rather than redesigned into hidden‑state (RNN/SSM) models through targeted fine‑tuning or distillation}.  These methods preserve most of the teacher's parameters while replacing quadratic self‑attention with sub‑quadratic mixers that maintain a \emph{single} recurrent state, thereby inheriting constant‑memory inference.

\paragraph{Cross‑architecture distillation.}  Earlier ``Transformer‑to‑RNN'' (T2R) conversions replaced softmax with trainable linear kernels but required heavy retraining.  SUPRA~\citep{mercat2024supra} refines this idea: starting from strong Llama‑2/Mistral checkpoints, it swaps attention for GroupNorm‑stabilized linear kernels and fine‑tunes on $\sim$20 B tokens, reaching competitive accuracy with only 5\% of the cost of pretraining a recurrent model from scratch. MOHAWK~\citep{bick2024mohawk} introduces a three‑phase procedure (matrix–orientation  hidden‑state alignment  knowledge distillation) that transfers a pretrained Transformer into a Mamba‑2 state‑space model using only 3B tokens, yielding ``Phi‑Mamba'' which outperforms all prior open recurrent LMs of similar size.  The same recipe scales to 1–8 B models in Llamba~\citep{bick2025llamba}, demonstrating that recurrent students can match Llama‑3 teachers with 0.1\% of original training compute while enabling larger batch sizes and higher.

\paragraph{Low‑rank linearization.}  LoLCATs~\citep{zhang2024lolcats} shows that high‑fidelity conversion does not need full‑model updates.  It first \emph{matches} every attention head with a sliding‑window linear mixer (attention transfer), then restores any residual loss with LoRA adapters touching just $0.2\%$ of weights.  This two‑step "low‑rank linearization" narrows the MMLU gap to $\le$1\% for 8 B models and scales to 70–405 B parameters within a single day of training.

\paragraph{Gated conversions.}  Liger~\citep{lan2025liger} repurposes the \emph{pretrained} key matrix to build per‑channel forget gates, yielding a gated recurrent student that recovers 93\% of teacher performance with only $0.02\%$ of the original token budget and no extra parameters beyond LoRA.

\section{Mechanistic Interpretability}
\label{sec:interpretability}
This section demonstrates the feasibility of Latent CoT and justifies the use of layers as indicators to facilitate the implementation of Latent CoT.
As previously discussed, the majority of latent reasoning behaviors in large language models emerge through operations across layers, both in temporal and spatial dimensions. This raises a fundamental question: \textbf{Are layers the basic computational units of reasoning? }

Mechanistic Interpretability providing tools like Probing and Circuit Analysis, enables us to shift from observing model behavior in reasoning to understanding its mechanism. This is crucial to unveil the role of Transformer’s layers in reasoning.
In this section, we first summarize existing work from an interpretability perspective to address \textbf{whether layer stacking represents a form of Latent CoT}.
Next, we analyze \textbf{how layers function as a latent CoT} by examining aspects such as layer specialization and inter-layer information flow.
Finally, we illustrate \textbf{the limitations of expressing CoT using layer representations}.






\subsection{Do Layer Stacks Reflect Latent CoT?}

The concept of Chain of Thought (CoT) reasoning allows models to generate sequential thought tokens, giving them more time and computational resources before arriving at an answer. This idea has been influential in shaping new paradigms for scaling inference in “thinking” models, such as OpenAI o1~\citep{jaech2024o1} and DeepSeek’s R1 \citep{guo2025deepseek}. In parallel, there’s growing evidence suggesting that the stacking of layers in neural networks similarly impacts reasoning capabilities, indicating a “layer-based hidden CoT.” This relationship between layer depth and latent reasoning is critical for understanding the model’s potential reasoning ability.

At a macro level, a series of studies have found \textbf{a close correlation between layer depth and the reasoning capabilities of the model}. \citet{yu2024llms} found that the model’s Implicit CoT capabilities are strictly limited by the number of network layers. For a 5-step reasoning task, although intermediate results emerge within some layers, the final reasoning outcome fails to emerge due to an insufficient number of layers. \citet{guo2025llms} discovered that at least 2-3 layers are required to form a complete two-step reasoning chain within the model. Insufficient layers or inadequate depth in subsequent layers will hinder the ability to perform multi-hop reasoning. In addition, some studies have explored the structural advantages brought by layer depth from the perspective of representational capacity. \citet{saunshi2025reasoning} formally establish that any K-layer transformer performing m-step CoT reasoning can be simulated by an (L+O(1)) layer transformer through m iterative forward passes. \citet{merrill2025little} demonstrate that increasing Transformer depth significantly enhances reasoning abilities, enabling complex tasks like language recognition and graph connectivity that fixed depths cannot achieve.
This theorem fundamentally establishes that \textbf{layer depth serves as the primary bottleneck for latent reasoning capacity}, where the achievable CoT step length scales linearly with layer count.

At a micro level, studies commonly reveal \textbf{a clear correspondence between specific layers and tasks within CoT reasoning}. Just like the various steps in CoT, different layers play distinct roles in the reasoning process, while the overall reasoning depth (layer count) influences the final reasoning performance. A series of interpretability studies have revealed significant functional differentiation across layers of varying depths in reasoning tasks ~\citep{hou2023towards, cabannes2024iteration}. Layer depth affects the completeness of reasoning chains~\citep{guo2025llms}, which expand in parallel and grow exponentially~\citep{wang2024towards}, with intermediate information being integrated and transmitted across depths~\citep{yu2025back}. These observations at the micro-level strongly suggest a structured functional differentiation across layers, each performing distinct computational roles analogous to steps in an explicit CoT. To better understand how this latent chain emerges from layer stacks, it is necessary to delve deeper into the specific mechanisms of layer specialization and inter-layer information flow.

\subsection{Mechanisms of Latent CoT in Layer Representation}
Following the evidence from the micro-level analysis, we formalize the theory of \textbf{Layer Specialization} as a foundational framework for interpreting Latent CoT. This perspective posits that individual layers within Transformer models systematically specialize to support distinct reasoning operations, collectively forming an implicit computational pipeline analogous to an explicit CoT. Next, we articulate the role each layer group (shallow, intermediate, and deep) plays in supporting this latent reasoning structure, followed by a discussion of how information is propagated across these specialized layers.
\paragraph{Theory of Layer Specilization}
The Transformer model consists of alternating self-attention and feed-forward network (FFN) modules. A natural assumption is that different layers play distinct roles in reasoning tasks~\citep{zhang2024investigating, gromov2024unreasonable, shi2024understanding}. A series of interpretability studies are focusing on uncovering how these layers work together to build and convey the underlying CoT processes. From shallow to deep layers, the model exhibits a clear “division of labor.” The reasoning process transitions from specific, local, and syntactic information in the shallow layers to rich semantic integration and the merging of reasoning paths in the intermediate and deep layers. This differentiated structure leads us to consider each layer as the smallest functional unit in the reasoning process.

\textbf{Shallow Layers: Basic representational processor of Latent CoT.} Transformer’s shallow layers perform initial text processing, laying the groundwork for higher-level semantic analysis and reasoning. 
Functionally, the shallow layers primarily process local information, syntactic structures~\citep{klein2022micse}, and surface patterns~\citep{geva2020transformer}, perform initial data transformations~\citep{chen2024unveiling}, and form early circuit primitives~\citep{wang2024loss, li2019enhancing, walkowiak2025unpacking}. Additionally, studies indicate that shallow layers are responsible for storing and recalling factual knowledge~\citep{yang2024large, skean2025layer} and bridging entity parsing in multi-hop reasoning tasks~\citep{yang2024large, shalev2024distributional, biran2024hopping}. 
In summary, shallow layers are crucial for processing fundamental information and factual knowledge, with their ability to establish bridging variables directly influencing the model’s reasoning performance.



\textbf{Intermediate Layers: Core of Latent CoT.} 
Intermediate layers play a pivotal role in complex, multi-step reasoning tasks for the following reasons: (1) Intermediate layers form specialized sub-circuits dedicated to reasoning functions, (2) Intermediate layers exhibit superior representational capabilities, and (3) Activations in intermediate layers have a decisive impact on reasoning outcomes.

\textbf{Intermediate layers contain specific, identifiable computational sub-circuits specialized for distinct reasoning sub-tasks}. These circuits typically involve coordinated interactions between attention heads and MLP modules. 
\citet{wang2022interpretability} reverse-engineer the internal algorithm by which GPT-2 identifies indirect objects in sentences. They identify a mid-layer attention sub-circuit responsible for entity tracking and pronoun resolution, showing that intermediate layers carry out essential structured reasoning. Similarly, a series of studies have identified potential reasoning circuits within the intermediate layers~\citep{hanna2023does, skean2024does,wang2022interpretability, wu2024unifying}. The formation of these circuits is emergent, representing efficient computational patterns spontaneously learned by the model from large-scale data~\citep{berti2025emergent, tang2025explainable}.

\textbf{Intermediate layers exhibit unique characteristics in representation}, not only demonstrating powerful expressive capabilities but also playing a crucial role in knowledge storage and encoding.
The performance of intermediate layer embeddings can exceed that of final layer embeddings by up to 16\% in text embedding tasks, and show consistency across different model architectures and scales~\citep{skean2025layer}.
Some researchers believe that this powerful representation capability stems from the objective function used during pretraining. The autoregressive paradigm induces an information bottleneck at intermediate depths of the model, forcing it to distill the most essential and salient information~\citep{lei2025representation, wang2022interpretability}.

\textbf{Intermediate layers have a causal influence on final reasoning outcomes}. Correct activation of these layers is necessary for the model to produce valid inferences. 
A series of studies identify specialized neurons in intermediate layers and perform causal interventions. They find that enhancing activations significantly improves reasoning performance, while suppressing activations leads to a decline in reasoning ability~\citep{wang2024unveiling, geiping2025scaling}.
Intermediate layer representations, acting as bridging entities, also play a causally critical role in multi-step reasoning outcomes~\citep{yang2024large}. The functional specialization of intermediate layers makes their correct activation critically decisive for the final reasoning outcomes. For example,  Ref.~\cite{li2024understanding} traced failures in multi-hop reasoning to specific Attention modules in the intermediate layers that improperly handled implicit reasoning steps. By successfully "patching" these modules to correct the reasoning, they provided strong causal evidence for the functional specialization of these intermediate layer circuits.

\textbf{Deep Layers: Output Refinement and Decision-making of Latent CoT.}
The deep layers of Transformer models lie at the end of the information processing flow, play a pivotal role in \textbf{output optimization and decision-making}. Deep layers receive rich representational information from intermediate layers and perform semantic transformation tailored to specific downstream tasks ~\citep{lei2025representation, skean2025layer}, performing more complex logical integration and determine the final answer~\citep{hanna2023does, dutta2024think}. 

However, several layer pruning studies indicate that deeper layers exhibit characteristics such as \textbf{poor training performance, limited functionality, and reduced representation learning capabilities}~\citep{yuan2024lift, csordas2025language, shemiranifar2025void}. Existing research attributes this degradation to variance issues in Pre-Layer Normalization and the frequent degeneration of attention matrices.~\citet{sun2025curse} suggest that the exponential growth of output variance in Pre-LN and derivatives approaching the identity matrix in deeper layers are the main causes of layer degradation.~\citet{sanyal2024inheritune} found that attention matrices in deeper layers frequently degenerate, often collapsing into nearly rank-one single-column patterns. We believe that maintaining the “effectiveness” of each layer during pre-training is crucial. Enhancing the functionality of layers, especially deep layers, is  a future direction to improve the model’s reasoning abilities.




\paragraph{Theory of Information Flow}
Given the layer specialization, the flow of information across these layers is crucial for reasoning process. 
\citet{stolfo2023mechanistic} quantify the indirect contributions of MLP and attention modules to clarify internal information flow pathways in LLM during arithmetic tasks. The results highlight the crucial role of the attention mechanism in inter-layer information flow during reasoning, which transmits computational information from early processing layers to the final token.
\citet{wang2024grokked} discover a “generalizing circuit” emerging during the grokking process. This circuit enables cross-layer information flow, with lower layers extracting bridge entities and higher layers conducting reasoning.~\citet{yu2025back} present a neuron-level investigation into the logits flow of LLMs during multi-hop knowledge prediction. With "back attention" mechanism, hidden information can be effectively transmitted from higher layers to lower layers, enhancing model's reasoning ability. Further research substantiates this by analyzing the "embedding trajectory" across all model layers. One study~\cite{wang2024coe}, which terms this the "Chain-of-Embedding," shows that the trajectory's geometric shape can distinguish correct from incorrect answers, enabling output-free self-evaluation. Another study~\cite{wang2024embeddingtra} uses trajectory "volatility" to detect out-of-distribution mathematical problems, finding that models show an "Early Stabilization" in their reasoning path for familiar tasks but not for unfamiliar ones. Both studies confirm that the vertical, layer-by-layer processing of LLMs contains a rich, interpretable information flow analogous to a latent chain of thought.

\subsection{Turing Completeness of Layer-Based Latent CoT}
Turing completeness is a fundamental concept in theoretical computer science. It describes the ability of a system to perform any computation that can be performed by a universal Turing machine. A computational system is considered Turing complete if it can simulate the computational process of any Turing machine. In this section, we first attempt to answer \textbf{whether the Vanilla Transformer is Turing complete}. Next, we summarize \textbf{what modifications are needed to make the Transformer achieve Turing completeness}.

\paragraph{Proof of Turing completeness in model architectures} Before the emergence of Transformers ~\cite{vaswani2017attention}, Recurrent Neural Networks (RNN)~\cite{Jordan1986OutsiderView, Elman1990FindingStructure} were the dominant architecture for processing sequential data. Owing to their inherent recursive nature, RNNs were theoretically proven to be Turing complete as early as 1996, setting a precedent for neural networks to achieve universal computational capabilities~\cite{Siegelmann1995ComputationalPower}. Subsequently, LSTM~\cite{Hochreiter1997LSTM} and GRU~\cite{Cho2014LearningPhrase} were proposed to address the vanishing gradient problem in RNNs, enabling more stable memory states over long sequences. 

A series of research efforts have attempted to prove the Turing completeness of Transformers from an architectural perspective under certain assumed constraints.~\citet{perez2019turing} formally proved for the first time that \textbf{the Transformer architecture is Turing complete}, possessing the universal capability to execute any computable function. However, the validity of this proof relies on three crucial theoretical assumptions: Arbitrary Precision, Positional Encodings, and Hard-Max Attention. Following this idealized and groundbreaking proof, more researchers began to consider the conditions under which a Transformer can achieve Turing completeness.
Further, ~\citet{li2025constant} proved for the first time that Turing completeness can be achieved under constant numerical precision. This study directly addresses the controversial assumption of infinite precision from earlier proofs, bringing the theoretical model closer to the computational constraints of the real world.

\paragraph{Proof of Turing completeness with Chain-of-Thought}
Additionally, another research path focuses on achieving more universal computational capabilities through CoT reasoning. Functionally, CoT transforms the Transformer from a limited context window into a dynamic computational tape. The model employs an autoregressive approach, writing each step's calculation result on a notepad and reusing the intermediate results in subsequent calculations.~\citet{qiu2024ask} proposed that \textbf{“prompting is Turing complete”.} They demonstrate that a single, finite-sized Transformer, as long as it is given a suitably constructed prompt, can compute any computable function.This is the first time the Turing completeness of Transformers has been revealed from the perspective of prompts.~\citet{li2024chain} discovered that a Transformer with constant depth can simulate a Boolean circuit of size T, provided it is allowed to perform T-step CoT reasoning.
These studies on the Turing completeness of CoT indicate a shift in the definition of general computation. Generality does not necessarily need to be embedded within the model architecture; it can also be achieved through interaction paradigms using fixed-depth models. 

\paragraph{Enhancing Transformer for Turing Completeness}
Beyond theoretical proofs, a series of studies have enhanced the expressive power of Transformers through architectural modifications, aiming to approach their theoretical limit of Turing completeness. A series of studies have introduced recurrent mechanisms to break through the fixed depth constraints of Transformers, as discussed in Section~\ref{sec:latent cot reasoning}. Additionally, some studies have incorporated external memory into Transformers~\citep{bulatov2022recurrent}.

\paragraph{A Unifying View of Implicit and Explicit Reasoning}
The reasoning process of Transformers can be viewed as “thought unfolding” across two dimensions. The well-known CoT unfolds along the “horizontal” sequence dimension, creating visible reasoning steps. Meanwhile, the network’s layer-by-layer computation can be seen as implicit unfolding and refinement of each token along the “vertical” depth dimension. 
As discussed above, CoT acts as the scratchpad between questions and answers, allowing the model to perform reasoning in an auto-regression mode, theoretically possessing Turing completeness. 
Meanwhile, each layer of the Transformer represents an implicit reasoning step, progressively optimizing the prediction of the next token.
\textbf{Thus, both methods represent a form of computational expansion, differing fundamentally in whether they unfold across the sequence or through the network’s depth.}

\begin{table*}[ht]
\centering
\renewcommand\arraystretch{1.2}
\begin{tabularx}{\textwidth}{@{} >{\raggedright\arraybackslash}p{0.16\textwidth} *{6}{>{\raggedright\arraybackslash}X} @{}}
\toprule
\textbf{} & 
\textbf{Operation} & 
\textbf{Storage} & 
\textbf{Resource Constraint} & 
\textbf{Optimization Objective} \\
\midrule

\textbf{Standard CoT} & 
Full Model Forward Pass & 
Explicit Tokens In the Sequence & 
Context Window & 
End-to-end Task  \\

\textbf{Layer-based Latent CoT} & 
Single Layer Forward Pass & 
Hidden States & 
Layer Nums & 
Next Token Prediction  \\

\bottomrule
\end{tabularx}
\caption{A comparison of Standard Chain-of-Thought (Horizontal Expansion) and Layer-based Latent CoT (Vertical Expansion) across key computational dimensions. 
}
\label{tab:cot_comparison_en}
\end{table*}

Moreover, a series of studies have sought to break the boundary between implicit CoT and explicit CoT. ~\citet{chowdhury2024investigating} propose Universal Transformers (UTs), which approach Turing completeness by implementing adaptive computation depth. The core idea of UTs is to repeatedly apply the same Transformer block across multiple “layers” or computational steps, thereby introducing a form of recurrence into the architecture.~\citet{zelikman2024quiet} integrate the CoT between layers and CoT between tokens, allowing for the output of intermediate thought processes among tokens as well. Furthermore, they proposed Fast Quiet-Star, which retains the token-level thinking trace while reducing computational cost.~\citet{dong2025reinforcement} reframed next-token prediction as a reasoning task trained using Reinforcement learning, where the model receives verifiable rewards for correctly predicting the next token for a given context.

\section{Towards Infinite-depth Reasoning}
\label{sec:infinite-depth reasoning}

Infinite-depth reasoning refers to an AI’s ability to devote unbounded “thinking time” to refine and perfect a solution irrespective of output length. In this section we first introduce spatial infinite-depth reasoning and then temporal infinite-depth reasoning. Spatial infinite-depth reasoning is realized by diffusion models that begin with a fully masked or noisy draft of the entire output and iteratively denoise it in parallel: each pass has bidirectional access to the full context, enabling global planning, logical consistency across distant segments, and iterative self-correction, with the number of refinement steps adjustable at inference time to trade speed for depth of reasoning. Temporal infinite-depth reasoning, by contrast, relies on autoregressive extensions that generate tokens one at a time in a left-to-right stream and can in principle produce arbitrarily long sequences—but their irreversible early decisions can accumulate errors and limit true global coherence.

\begin{table*}[t!]
\centering
\renewcommand\arraystretch{1.2}
\begin{tabular}{@{}p{0.55\textwidth}>{\RaggedRight}p{0.4\textwidth}@{}}
\toprule
\textbf{Method} & \textbf{Unified Latent-Update Formula} \\
\midrule
\multicolumn{2}{l}{\textbf{Masked Diffusion Models (Temporal-only)}} \\
\addlinespace[0.2em]
\quad $\bullet$ D3PM~\citep{austin2021structured}, SEDD~\citep{lou2024discrete}, RADD~\citep{ou2024your} & \multirow{3}{0.53\textwidth}
{
$\bm x_{t+1}^{l}(i) = f(\bm x_t^l(i))$
} \\
\quad $\bullet$ MD4~\citep{shi2024simplified}, Simple-MDM~\citep{sahoo2024simple}, MDM~\citep{nie2024scaling} & \\
\quad $\bullet$ MMaDA~\citep{yang2025mmada}, IRED~\citep{Du_2024_ICML} & \\
\addlinespace[0.3em]
\midrule
\multicolumn{2}{l}{\textbf{Masked Diffusion Models (With Cache)}} \\
\addlinespace[0.2em]
\quad $\bullet$ LLaDA~\citep{nie2025large}, dKV-Cache~\citep{ma2025dkv}, DoT-SEDD~\citep{ye2024diffusionthoughtschainofthoughtreasoning} & \multirow{3}{0.53\textwidth}
{\parbox{0.51\textwidth}{
$\bm x_t^{l+1} = f_{\tau}(\bm x_t^l, \mathbf{S}_t^l)$\\[0.3em]
$\mathbf{S}_{t+1}^l(i) \approx g_{\tau}(\bm x_t^l(i), \mathbf{S}_t^l(i)) $
}} \\
\quad $\bullet$ dLLM-Cache~\citep{liu2025dllmcacheacceleratingdiffusionlarge}, d1-LLaDA~\citep{zhao2025d1scalingreasoningdiffusion}, DCoLT~\citep{huang2025reinforcingdiffusionchainlateral} & \\
\quad $\bullet$ LLaDA~1.5~\citep{zhu2025llada15variancereducedpreference}, MGDM~\citep{ye2025autoregressiondiscretediffusioncomplex} & \\
\addlinespace[0.3em]
\midrule
\multicolumn{2}{l}{\textbf{Embedding-based Diffusion Models}} \\
\addlinespace[0.2em]
\quad $\bullet$ Diffusion-LM~\citep{li2022diffusionlmimprovescontrollabletext}, CDCD~\citep{dieleman2022continuousdiffusioncategoricaldata}  & \multirow{3}{0.53\textwidth}
{\parbox{0.51\textwidth}{
$\bm x_{t+1}^l = f(\bm x_t^l, \epsilon_t)$
}}\\
\quad $\bullet$ Plaid~\citep{gulrajani2023likelihoodbaseddiffusionlanguagemodels}, DoT-Plaid~\citep{ye2024diffusionthoughtschainofthoughtreasoning} & \\
\quad $\bullet$ TESS~\citep{mahabadi2024tesstexttotextselfconditionedsimplex}, TESS 2~\citep{tae2025tess}, Bit Diffusion~\citep{chen2023analogbitsgeneratingdiscrete} & \\
\addlinespace[0.3em]
\midrule
\multicolumn{2}{l}{\textbf{Hybrid AR-Diffusion Models}} \\
\addlinespace[0.2em] 
\quad $\bullet$ DiffuLLaMA~\citep{gong2025scaling}, Dream~\citep{dream2025} & \multirow{3}{0.53\textwidth}{\parbox{0.51\textwidth}{
$\bm x_t^{l+1} = f_{\tau}(\bm x_t^l, \mathbf{S}_t^l)$\\[0.3em]
$\mathbf{S}_{t+1}^l(i) = g_{\tau}(\bm x_t^l(i), \mathbf{S}_t^l(i)) $\\[0.2em]
+ AR prefix caching
}}\\
\quad $\bullet$ L2D~\citep{cetin2025large}, Gemini Diffusion~\citep{google2025geminiDiffusion} & \\
\quad $\bullet$ Mercury~\citep{labs2025mercuryultrafastlanguagemodels} & \\
\addlinespace[0.2em]
\bottomrule
\end{tabular}
\caption{Text diffusion models organized by cache integration capabilities, showing the evolution from temporal-only updates to spatial-temporal frameworks with KV cache mechanisms.}
\label{tab:enhanced_diffusion_models}
\end{table*}


\begin{figure}[t!]
    \centering
    \includegraphics[width=0.9\linewidth]{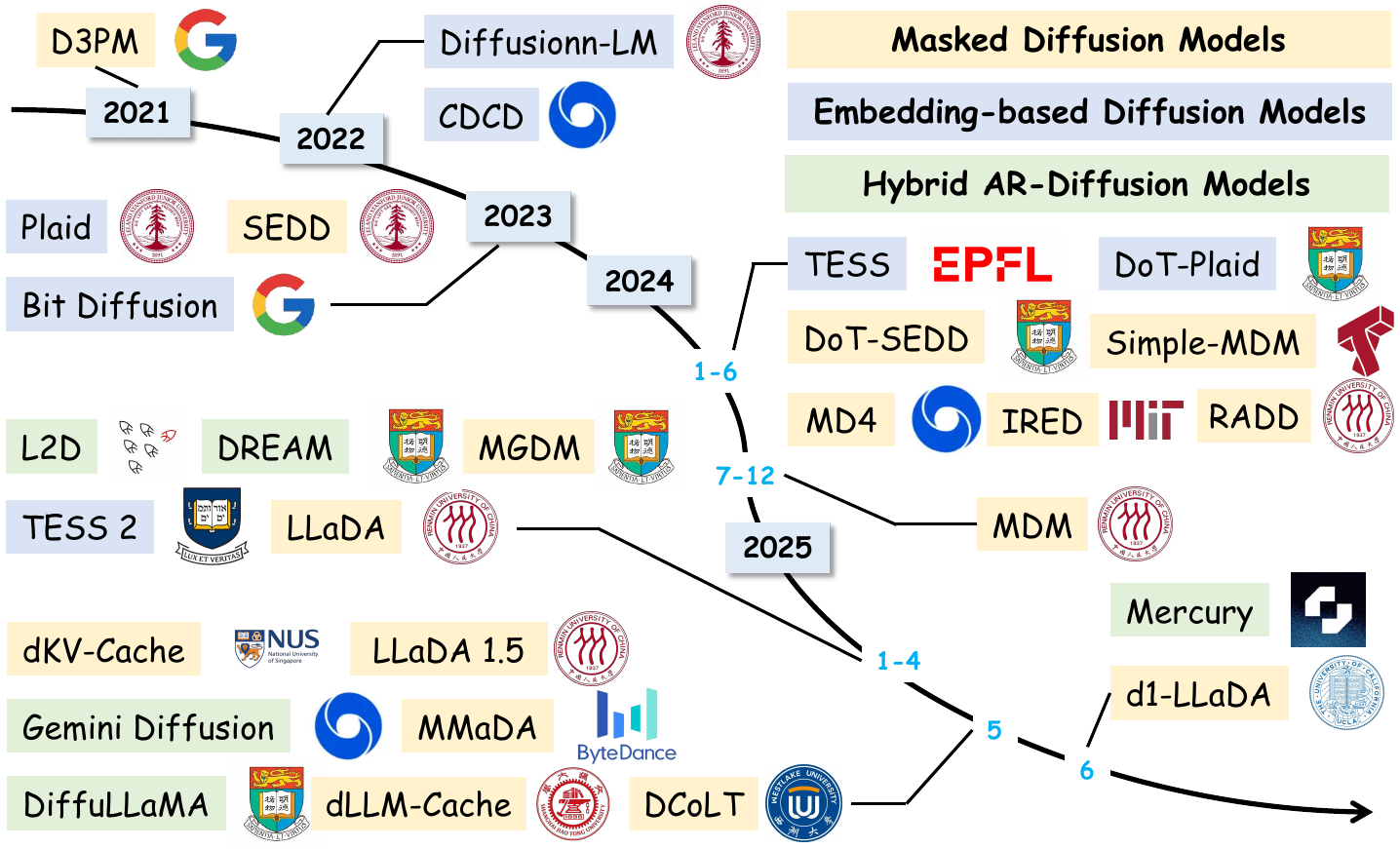}
    \caption{An evolutionary graph of the text diffusion models, including three architectural families: Masked Diffusion Models, Embedding-based Diffusion Models, and Hybrid AR-Diffusion Models.}
    \label{fig:evol_graph_diffusion}
\end{figure}

\subsection{Spatial Infinite Reasoning: Text Diffusion Models}

Text diffusion models represent a paradigm shift for complex reasoning tasks, offering an alternative to traditional AR generation. Unlike sequential models that generate text token-by-token, diffusion models enable \emph{spatial infinite reasoning} through iterative global refinement. This approach allows models to engage in holistic planning and develop logical connections across the entire reasoning chain simultaneously, overcoming the limitations of sequential generation where early decisions become irreversible constraints.
The connection between diffusion models and infinite depth reasoning lies in their iterative refinement capacity. While traditional models are constrained by fixed computational depth, diffusion models can theoretically refine reasoning through unlimited denoising steps. Each step provides additional reasoning depth, allowing progressive elaboration from high-level plans to detailed solutions. 

We organize text diffusion models into three architectural families: \emph{Masked Diffusion Models} that enable bidirectional context awareness, \emph{Embedding-based Diffusion Models} that preserve structured reasoning while enabling global refinement, and \emph{Hybrid AR-Diffusion Models} that combine diffusion and AR paradigms.
 
\subsubsection{Masked Diffusion Models}

Masked Diffusion Models (MDMs) exemplify spatial reasoning in text generation. These models operate on complete text sequences where tokens are initially masked, requiring simultaneous prediction of all missing tokens based on bidirectional context. This provides full access to the entire information landscape at each denoising step.

\textbf{MDMs adopt a latent update mechanism driven by an explicit token-level mask $M_t$ at each step $t$}, the corresponding unified latent-update formulas are described in Table~\ref{tab:enhanced_diffusion_models}.
For \textbf{temporal-only MDMs}, the latent update formula, $\bm x_{t+1}^{l}(i) = f(\bm x_t^l(i))$, describes how the representation of an individual token, $\bm x_{t}^{l}(i)$, is updated from denoising step $t$ to $t+1$ within a specific layer $l$. This indicates a direct token-level update, where the model's focus is on iteratively refining the masked parts of the sequence.
For \textbf{MDMs with cache}, two formulas describe the process. The first formula, $\bm x_t^{l+1} = f_{\tau}(\bm x_t^l, \mathbf{S}_t^l)$, describes the temporal transformation. It shows that at denoising step $t$, the output $\bm x_t^{l+1}$ for layer $l+1$ is generated by a function $f_{\tau}$ that takes the current token representations $\bm x_t^l$ and the current KV-cache $\mathbf{S}_t^l$ as input. This indicates that the Transformer block's processing for generating token representations directly leverages the KV-cache for spatial context.


The iterative unmasking process enables sophisticated reasoning capabilities impossible in sequential generation. 
Some pioneers provide a strong foundation for masked diffusion models and supporting reasoning tasks through better input, intermediate steps, and outputs. 
D3PM~\citep{austin2021structured} goes beyond corruption processes with uniform transition probabilities, while SEDD~\citep{lou2024discrete} introduces the EBLO loss that naturally extends score matching to discrete spaces. 
Further refinements, like RADD~\citep{ou2024your}, MD4~\citep{shi2024simplified}, and Simple-MDM~\citep{sahoo2024simple},  have streamlined training through hybrid masked losses, facilitating conversion of encoder models like BERT into effective generative reasoning systems. 
Besides, MMaDA~\citep{yang2025mmada} adopts a unified diffusion architecture for multi-modal reasoning and aligns reasoning processes between textual and visual domains.
Research has shown that MDMs can be scaled effectively, achieving strong performance and efficiency~\citep{nie2024scaling}. 
IRED~\citep{Du_2024_ICML} framed reasoning as an energy minimization process implemented through diffusion models. This approach enables iterative refinement from vague reasoning paths to precise solutions, particularly effective for multi-constraint problems. Energy diffusion demonstrated significant advantages over traditional methods in complex reasoning tasks.
%
%
The LLaDA model~\citep{nie2025large} uses discrete random masking, enabling sophisticated capabilities like reverse-order reasoning.
To accelerate masked diffusion language models, dKV-Cache~\citep{ma2025dkv} introduces a delayed and conditioned key–value caching strategy that achieving up to 2-10$\times$ inference speedup. dLLM-Cache~\citep{liu2025dllmcacheacceleratingdiffusionlarge} introduces an adaptive caching strategy achieves up to 9.1$\times$ speedup over the standard inference method of LLaDA~\citep{nie2025large}.
DoT-SEDD \citep{ye2024diffusionthoughtschainofthoughtreasoning} subsequently generalized chain‑of‑thought (CoT) reasoning to the MDM framework, enhancing coherence and accuracy through natural self-correction, with particular strengths in mathematical reasoning.
The framework has been extended through Multi-Granularity Diffusion Modeling (MGDM)~\citep{ye2025autoregressiondiscretediffusioncomplex}, which prioritizes difficult subgoals and achieves state-of-the-art results on complex planning tasks.
d1-LLaDA~\citep{zhao2025d1scalingreasoningdiffusion} introduces diffu‑GRPO, a lightweight policy‑gradient algorithm tailored to masked diffusion models that surpasses SFT across mathematical and planning benchmarks. LLaDA~1.5~\citep{zhu2025llada15variancereducedpreference} advances this line with VRPO, which combines unbiased Monte‑Carlo budget allocation and antithetic sampling to sharply reduce the variance of ELBO‑based preference optimization. DCoLT~\citep{huang2025reinforcingdiffusionchainlateral} applies outcome‑based reinforcement learning by using a probabilistic policy or ranking‑based Unmasking Policy Module to jointly optimize the entire reasoning trajectory.

\subsubsection{Embedding‑based Diffusion Models}

Embedding‑based diffusion models (EDMs) extend the paradigm of spatial reasoning by first mapping discrete token sequences into continuous token embeddings and then operating on these embeddings, where they are disrupted with Gaussian noise. The models denoise every latent vector using bidirectional context, enjoying complete visibility of the information landscape at each refinement step. Although this high‑level objective mirrors that of MDMs, EDMs inhabit a fundamentally different design space due to their continuous embeddings formulation.

\textbf{EDMs achieve latent update by applying noise to all tokens uniformly and allow denoising dynamics to determine recovery,} the corresponding unified latent-update formulas are described in Table~\ref{tab:enhanced_diffusion_models}.
The formula describes how the representation of the entire sequence's tokens, $\bm x_t^l$, is updated from denoising step $t$ to $t+1$ for a given layer $l$, enabling iterative refinement within the continuous latent space. The function $f$ represents the diffusion model's core denoising network (typically a Transformer), taking the current noisy embeddings $\bm x_t^l$ and a noise term $\epsilon_t$ to compute the denoised embeddings. Conceptually, this process operates on the entire sequence's embedding representation, rather than specific parts of individual tokens or their hidden states.

Early EDM research emphasized controllable generation~\citep{li2022diffusionlmimprovescontrollabletext} and sequence-to-sequence tasks~\citep{dieleman2022continuousdiffusioncategoricaldata, mahabadi2024tesstexttotextselfconditionedsimplex}, as well as efficient latent encodings of discrete sequences~\citep{chen2023analogbitsgeneratingdiscrete, gulrajani2023likelihoodbaseddiffusionlanguagemodels, tae2025tess}. Plaid~\citep{gulrajani2023likelihoodbaseddiffusionlanguagemodels} systematically characterizes the capacity of this model family by deriving empirical scaling laws, closing the compute‑efficiency gap with autoregressive language models to 64$\times$. DoT-Plaid \citep{ye2024diffusionthoughtschainofthoughtreasoning} subsequently generalized chain‑of‑thought (CoT) reasoning to the EDM framework, allowing entire reasoning paths to evolve through iterative latent refinement and enhancing coherence and accuracy through natural self-correction, with particular strengths in mathematical reasoning.

\subsubsection{Hybrid AR-Diffusion Models}

The third family explores direct integration of diffusion and autoregressive paradigms, creating hybrid systems that leverage complementary strengths. These models recognize that while diffusion excels at global planning, autoregressive generation remains effective for certain sequential dependencies.

\textbf{Hybrid AR-Diffusion models integrate autoregressive generation with diffusion-based latent refinement, combining the strengths of sequential coherence and bidirectional reasoning.} The corresponding unified latent-update formulas are described in Table~\ref{tab:enhanced_diffusion_models}.
The formula $\bm x_t^{l+1} = f_{\tau}(\bm x_t^l, \mathbf{S}_t^l)$ details the temporal transformation. Here, a Transformer block $f_{\tau}$ refines token representations $\bm x_t^l$ from layer $l$ to $l+1$ at denoising step $t$. This refinement explicitly uses the current KV-cache $\mathbf{S}_t^l$. Crucially, this temporal update is enhanced by \textbf{AR prefix caching}, which brings in forward-context alignment from the already generated text $x_{<t}$.
The second formula governs the spatial update of the KV-cache for individual token $i$. This update is driven by the function $g_{\tau}$ that takes the token's representation $\bm x_t^l(i)$ and its old cache $\mathbf{S}_t^l(i)$ as input. The explicit inclusion of ``AR prefix caching'' in this formula indicates that the KV-cache update directly incorporates AR prefix caching mechanisms, enhancing the cache with forward context. This allows the model to dynamically stabilize reliable representations, focusing refinement on uncertain tokens while leveraging the strength of pre-existing sequential information.

DiffuLLaMA~\citep{gong2025scaling} introduces a continual pre-training approach that converts existing autoregressive models (like GPT-2 and LLaMA) into diffusion models, which provides a powerful and scalable tool for complex reasoning tasks that demand efficient and flexible text processing.
L2D~\citep{cetin2025large} uses a modular design integrating a diffusion pipeline with a pre-trained autoregressive model, creating synergy between global reasoning and sequential fluency. The Dream model~\citep{dream2025} leverages autoregressive initialization for training stability and context-adaptive noise scheduling. 
By leveraging a diffusion method for parallel, coarse-to-fine token generation, commercial frameworks such as Gemini Diffusion~\citep{google2025geminiDiffusion} and Mercury~\citep{labs2025mercuryultrafastlanguagemodels} significantly boost the speed and efficiency of code processing in large language models. This provides a more effective solution for latency-sensitive reasoning tasks like chain-of-thought and agentic workloads.
These hybrid approaches represent a promising direction, acknowledging that different reasoning aspects may benefit from different computational paradigms.

\subsection{The optimization-Based Perspective: Trading Time for Depth}
The optimization-based perspective introduced in~Section \ref{sec:opt-state-evo} suggests that \textbf{time itself can be traded for network depth}. When the hidden state $\mathbf{S}_t$ is updated by a gradient-like rule $\mathbf{S}_{t}=\mathbf{S}_{t-1}-\eta_t\nabla_{\mathbf{S}}\ell(\mathbf{S}_{t-1};\mathbf{k}_t,\mathbf{v}_t)$, each additional token performs one extra step of a (stochastic) optimizer that refines an implicit layer. Consequently, processing a longer sequence is mathematically equivalent to running the same layer for more optimization iterations, thereby yielding \textbf{greater reasoning depth without adding parameters}. This observation converts the long‑context challenge into a new question: \textbf{how can we instantiate a \emph{network of unbounded depth} that remains trainable and efficient?}

\subsubsection{Towards an 'Infinitely Long' Optimizer Network}
Recent work pursues three complementary strategies:

\paragraph{Infini‑attention:}~\citet{munkhdalai2024leave} attach a compressive memory to every Transformer block. Each incoming segment updates this memory via a linear--delta rule that asymptotically approaches the fixed‑point of an associative array, allowing the model to stream \emph{infinitely} long inputs with $\mathcal{O}(1)$ memory. However, a reproduction~\cite{huggingface_infini_attention_2024} attempt documented significant practical challenges with this approach. Their key finding was that the model's long-context performance degraded as the number of memory compression steps increased. The authors also reported severe convergence issues, particularly with the gating parameters that balance local and compressed memory, ultimately concluding that the technique was not reliable for extending pretrained models. This empirical evidence suggests that while the idea of memory compression is promising, the specific mechanism in Infini-attention may not be effective in practice, and other methods like rope scaling or Ring Attention are currently more viable options.

\paragraph{Test‑time training (TTT) and its descendants:} ~\citet{sun2024ttt} pioneered the idea of performing a few steps of SGD on the hidden state during inference. Follow‑up models like Titans~\citep{behrouz2024titans}, OmegaNet and Atlas~\citep{behrouz2025atlas}, replace first‑order updates with Adam- or Muon‑style optimizers and introduce chunk‑wise parallelism so that $10^6$–token streams can be handled on modern accelerators. Empirically, Titans‑S ($\!\sim$250\,M) already matches a 1.3\,B Transformer on 1‑shot recall after only $\sim$1\,M optimization steps, demonstrating that ``deeper through time'' can substitute for ``deeper via layers''.

In contrast to methods relying on frequent, small-batch updates, recent work~\cite{zhang2025test} argues that this strategy suffers from severe computational inefficiency due to low hardware utilization. The proposed solution, Large Chunk Test-Time Training (LaCT), advocates for the opposite: updating "fast weights" using extremely large chunks of data, ranging from thousands to over a million tokens. This large-chunk paradigm dramatically improves hardware utilization without custom kernels and, more importantly, enables the scaling of nonlinear state sizes to a much larger fraction of the model's parameters (up to 40\%). This enhanced state capacity, combined with sophisticated optimizers like Muon, has been validated across diverse tasks, including novel view synthesis, language modeling, and autoregressive video diffusion.

\paragraph{Implicit Fixed‑Point RNNs:}
An orthogonal line of work revisits classical RNNs through the lens of \emph{implicit layers}.~\citet{schone2025implicit} show that iterating a state‑space block until convergence yields non‑linear, non‑diagonal transitions that recover the expressivity of general RNNs while retaining training parallelism. Practically, one runs only a small, adaptive number of self‑iterations ($\leqslant 16$ for most natural‑language tokens), giving another route to unbounded depth: the model simply halts when additional refinement becomes irrelevant.

\subsubsection{A Unifying View}
All three families embody the same principle:
\begin{center}
    \emph{\textbf{Depth emerges from optimization over time.}}
\end{center}
The hidden state plays the role of a ``fast‑weight'' layer whose parameters are refined either \textbf{explicitly} (TTT, Titans, Atlas), \textbf{implicitly} (fixed‑point RNNs), or through an \textbf{associative cache} (Infini‑attention). Longer sequences therefore unlock deeper reasoning. Crucially, chunk‑wise scans and parallel fixed‑point solvers keep the wall‑clock cost nearly linear, enabling experiments with \textbf{million‑token} contexts on a single GPU.

\section{Discussion and Conclusion}
This survey provides a comprehensive overview of Latent CoT and reasoning, an emerging paradigm in AI reasoning. While large language models have demonstrated impressive reasoning using explicit CoT that verbalizes intermediate steps, this approach is limited by the expressive bandwidth of natural language. Latent CoT addresses this by shifting the entire reasoning process into the model's continuous hidden state, aiming to enhance expressive power and performance. By operating in a continuous space, the model is freed from the constraints of a finite token vocabulary and can explore more efficient and powerful reasoning strategies that may not have direct linguistic equivalents.

Latent reasoning methodologies primarily follow two paradigms: vertical and horizontal recurrence. Vertical recurrence, or activation-based methods, expands computational depth by iteratively refining information within the same set of layers, either through explicit architectural loops or induced through specialized training. In contrast, horizontal recurrence, or hidden-state-based methods, expands the model's temporal capacity by evolving a compressed hidden state over long sequences, allowing for the integration of vast amounts of information. These approaches are complemented by mechanistic interpretability research, which examines how different network layers specialize to form an implicit computational pipeline analogous to an explicit CoT.

Notably, this survey does not offer a direct empirical comparison across these varied models. The field is developing rapidly, with different models being created under disparate training conditions—some are pre-trained from scratch, while others are adapted from existing foundation models via continual pre-training. Furthermore, most studies compare their models to non-reasoning LLM baselines rather than to each other. This lack of consistent training methodologies and standardized benchmarks currently makes a direct, apple-to-apples comparison of empirical results challenging. It is our hope that a unified evaluation framework will emerge in the future to enable a clearer assessment of the relative strengths of these approaches.

The survey culminates by exploring the frontier of infinite-depth reasoning, which aims to give models the ability to use unbounded computational steps to refine a solution. Text diffusion models are a key innovation in this area, as they operate on the entire output sequence in parallel. This allows for global planning, iterative self-correction, and logically consistent reasoning processes that are not constrained by sequential, irreversible decisions. By unifying these perspectives, the survey charts the conceptual landscape of latent reasoning and points toward future directions in advanced AI cognition.







\newpage

\bibliography{main.bib}

\newpage

\end{document}